\documentclass[pdflatex,sn-nature]{sn-jnl}
\usepackage[utf8]{inputenc}

\usepackage{graphicx}
\usepackage{multirow}
\usepackage{amsmath,amssymb,amsfonts}
\usepackage{amsthm}
\usepackage{mathrsfs}
\usepackage[title]{appendix}
\usepackage{xcolor}
\usepackage{textcomp}
\usepackage{manyfoot}
\usepackage{booktabs}
\usepackage{bm}
\usepackage{mathtools}
\usepackage{graphicx}
\usepackage{subcaption}
\usepackage{placeins}
\usepackage{fancyhdr}

\theoremstyle{remark}
\newtheorem{remark}{Remark}

\newcommand{\cross}{\times}

\begin{document}
\newcommand{\shorttitle}{Physical neural networks with trainable nonlinear connections}
\newcommand{\maintitle}{Physical neural networks with trainable nonlinear connections for continuous control}

\title[\shorttitle]{\maintitle}

\author*[1]{\fnm{Ian T.} \sur{Vidamour}}\email{i.vidamour@sheffield.ac.uk}
\author[2]{\fnm{Fernando} \sur{Aguirre}}
\author[3]{\fnm{Thomas J.} \sur{Hayward}}
\author[1]{\fnm{Matthew O. A.} \sur{Ellis}}
\author[3]{\fnm{Charles} \sur{Swindells}}
\author[4]{\fnm{Alexander} \sur{McDonnell}}
\author[4]{\fnm{Martin} \sur{Trefzer}}
\author[1]{\fnm{Finley} \sur{Robins}}
\author[1]{\fnm{Luca} \sur{Manneschi}}
\author[5]{\fnm{Susan} \sur{Stepney}}
\author[6]{\fnm{Anthony J} \sur{Kenyon}}
\author[7, 8]{\fnm{Oliver J.} \sur{Sutton}}
\author[9, 10]{\fnm{Jack C.} \sur{Gartside}}
\author[7]{\fnm{Ivan Y.} \sur{Tyukin}}
\author[6]{\fnm{Adnan} \sur{Mehonic}}
\author*[1]{\fnm{Eleni} \sur{Vasilaki}}\email{e.vasilaki@sheffield.ac.uk}

\affil[1]{\orgdiv{School of Computer Science}, \orgname{University of Sheffield}, \orgaddress{\city{Sheffield}, \postcode{S1 4DP}, \country{United Kingdom}}}
\affil[2]{\orgname{Intrinsic Semiconductor Technologies}, \orgaddress{\city{Cambridge}, \country{United Kingdom}}, \postcode{CB4 0GJ}}
\affil[3]{\orgdiv{School of Chemical, Biological, and Materials Science Engineering}, \orgname{University of Sheffield}, \orgaddress{\city{Sheffield}, \postcode{S1 3JD}, \country{United Kingdom}}}
\affil[4]{\orgdiv{School of Physics, Engineering, and Technology}, \orgname{University of York}, \orgaddress{\city{York}, \postcode{YO10 5EZ}, \country{United Kingdom}}}
\affil[5]{\orgdiv{Department of Computer Science}, \orgname{University of York}, \orgaddress{\city{York}, \postcode{YO10 5EZ}, \country{United Kingdom}}} 
\affil[6]{\orgdiv{Department of Electronic \& Electrical Engineering}, \orgname{University College London}, \orgaddress{\street{Roberts Building, Torrington Place}, \city{London}, \postcode{WC1E 7JE}, \country{United Kingdom}}}
\affil[7]{\orgname{King's College London}, \orgaddress{\city{London}, \postcode{WC2R 2LS}, \country{United Kingdom}}}
\affil[8]{\orgname{Synoptix Ltd.}, \orgaddress{\city{Bristol}, \postcode{BS16 1EJ}, \country{United Kingdom}}}
\affil[9]{\orgdiv{Blackett Laboratory}, \orgname{Imperial College London}, \orgaddress{\city{London}, \postcode{SW7 2AZ}, \country{United Kingdom}}}
\affil[10]{\orgname{London Centre for Nanotechnology, Imperial College London}, \orgaddress{\city{London}, \postcode{WC1H 0AH}, \country{United Kingdom}}}

\abstract{Physical neural networks promise low-power machine learning by computing directly with analogue device physics, yet many architectures require nonlinear devices to act as scalar weights or as fixed, identical activation functions, discarding potential computational resources. Inspired by Kolmogorov--Arnold networks, we place a trainable nonlinear function on every connection, making each physical edge a learnable computational element. We realise these edges as analogue band-pass filters on field-programmable analogue arrays and transfer trained networks directly to hardware, with quantified fidelity across approximately $35{,}000$ experimentally realised connections. On smooth, continuously valued tasks learned from imperfect signals---six-axis robotic kinematics from actuator-corrupted observations, continuous-action reinforcement learning, and photovoltaic maximum-power-point tracking---the networks match or exceed multilayer perceptrons with substantially fewer trainable parameters and far fewer physical connections. A dedicated CMOS implementation of the photovoltaic controller is projected to consume approximately $35\;\mathrm{\mu W}$. Replacing the filters with simulated memristor circuits reproduces the advantage, showing that it follows from the architecture instead of one particular device. Making device nonlinearity trainable, instead of suppressing it, offers a route to compact analogue networks for the types of smooth, continuous inference and control tasks studied here.}

\keywords{physical neural networks, Kolmogorov--Arnold networks, analogue computing, low-power control, trainable nonlinear connections}

\maketitle
\section{Introduction} \label{sec:Introduction}
Physical neural networks implement inference directly in hardware, using the native response of analogue devices as computational primitives\cite{markovic_physics_2020,hermans_trainable_2015,nakajima2022physical,jaeger2023toward}. Implementations now span memristive\cite{jeong2016memristors,sebastian_memory_2020,aguirre2024hardware}, spintronic\cite{allwood2023perspective,stenning2024neuromorphic}, photonic\cite{kazanskiy2022optical}, and electronic\cite{liang2024physical} substrates, offering routes to low-power machine learning beyond conventional digital processors. However, most physical neural network architectures adopt a key assumption from conventional software-based neural networks: that connections between nodes are represented as scalar weights. In standard artificial networks, expressivity is achieved through dense linear connectivity between many nodes, each with a fixed, identical nonlinearity. When this architecture is mapped directly to analogue hardware, it can require the fabrication, programming, and control of many physical connections. At the same time, devices with rich nonlinear responses—such as a memristor's current–voltage curve or a filter's frequency response—may be limited to acting as programmable linear weights or fixed, identical activation functions.

Here we invert that assumption. Each connection between two network nodes, which we call an edge, carries a trainable nonlinear function rather than a scalar weight. As a result, the edge computes using the native response of the device, rather than merely scaling a signal. This architecture is related to the Kolmogorov--Arnold network\cite{liu2024kan}, a neural network model in which each connection is replaced by a learnable univariate function rather than a scalar weight; however, our approach is specifically motivated by a hardware principle: device nonlinearity should be used as the computational resource, not suppressed. We refer to this approach as a Physical Kolmogorov--Arnold-inspired Network (PhyKAN). Because each edge carries an expressive, smoothly tuneable function, we hypothesise that PhyKANs are especially suited to continuous control and regression—tasks that require learning smooth mappings over coupled real-valued variables and can therefore be represented with relatively few nodes and connections.

Prior work on physical neural networks has largely asked how to train these systems when a device's behaviour is only partially known: physics-aware training\cite{wright2022deep} and noise-aware dynamic optimisation\cite{manneschi2025noise} estimate gradients through a learned digital twin, while forward--forward and local-learning schemes\cite{momeni2023_PhyLL} remove the need for global backpropagation. Sharpness-aware training\cite{xu2026physical} improves robustness to model–reality mismatch. All these methods assume the conventional weight-and-activation architecture. In contrast, the complementary question—what architecture a physical system should implement in the first place—has received far less attention. Early explorations of trainable-edge architectures were limited to simulation\cite{peng2024photonic} or to small photonic modules using standard nonlinear components\cite{calccado2026small}. More recent demonstrations have shown that trainable edge nonlinearities can be built using compound memristor--transistor cells\cite{wen2026computing}, silicon-on-insulator synaptic nonlinear elements (SYNE)\cite{taglietti2026}, and reconfigurable nonlinear-processing units\cite{escudero2026physical}. Rather than asking whether such architectures can be built, here we investigate under what conditions the architectural advantages attributed to physical KAN-like networks---such as improved efficiency or expressivity---hold in continuous control and reinforcement-learning problems. These are precisely the tasks for which small-scale physical networks can feasibly be deployed in edge applications.

We address this question using analogue electronic filters as the physical edge, realised on Field-Programmable Analogue Arrays (FPAA). Each filter is set by two corner frequencies and a gain, which are tuned by gradient descent through its analytical transfer function. Banks of filters are summed to form a smooth univariate response within the filter bank's range with respect to frequency-encoded input. With sufficient filters, this basis can approximate any continuous one-dimensional response; at finite size, it favours smooth variation over sharp transitions. 

Across a range of tasks---including six-axis robotic kinematics learned from actuator-corrupted observations, continuous-action reinforcement learning with bootstrapped targets, and photovoltaic maximum-power-point tracking learned from environmental feedback---PhyKANs can match or exceed multilayer perceptrons with fewer trainable parameters, and therefore substantially fewer nodes and connections, on tasks requiring smooth, continuously valued mappings. In contrast, no consistent reduction in trainable parameters is observed on the Fashion-MNIST classification and binary-action CartPole benchmarks. Parameter-efficiency advantages occur most consistently for smooth, continuously valued mappings. Trained parameters transfer directly to hardware with quantified per-connection error, and the networks are projected to run at ${\sim}35\;\mathrm{\mu W}$ on a CMOS transconductor implementation. Finally, a mathematical analysis shows why individual physical edges can carry substantial approximation power: in the idealised setting, a single filter-based edge can approximate any continuous one-dimensional response, so smooth variation can be represented inside individual connections, and networks built from such edges can approximate continuous multivariate maps (Supplementary Note \ref{sec:uat}). A simulated memristor-based realisation with identical network architecture but different nonlinear basis functions shows similar parameter-efficiency advantages on the three continuous tasks, supporting the hypothesis that trainable nonlinear connections can provide the same architectural benefit across different physical substrates.

\section{Results}
\subsection{Constructing trainable nonlinear connections from analogue filters.}\label{sec:KANdefinition}
\begin{figure}[!htbp]
    \centering
    \includegraphics[width=\linewidth]{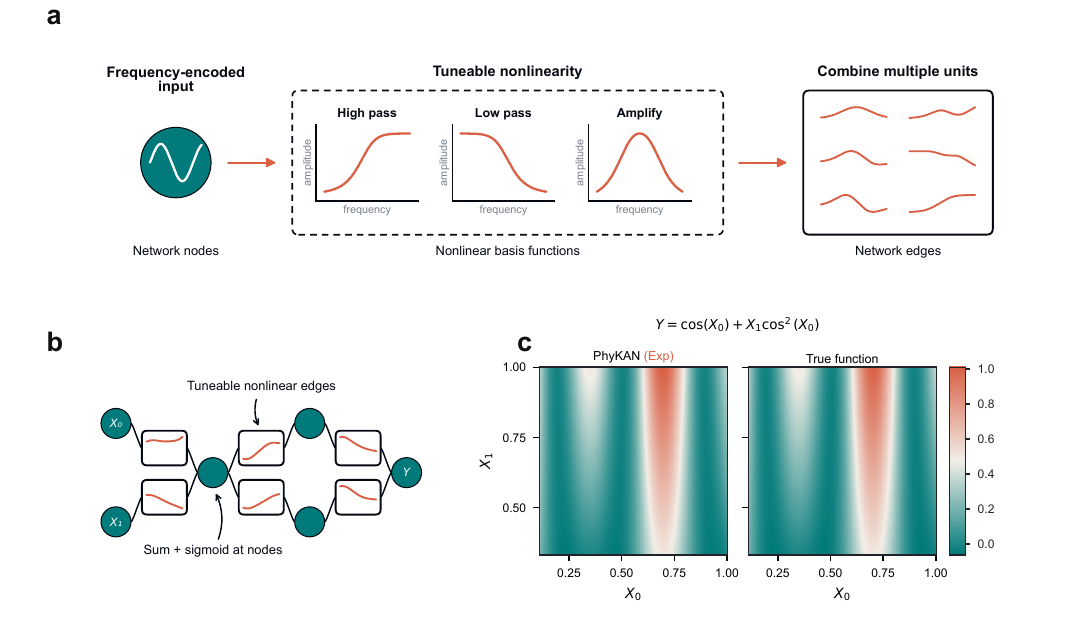}
    \caption[Constructing trainable nonlinear connections from analogue filters]{\textbf{Constructing trainable nonlinear connections from analogue filters.} \textbf{a}, Schematic of the components which make up a PhyKAN edge. Inputs are frequency-encoded at network nodes and passed to network edges. Each edge is a univariate function built from tuneable band-pass filters (high-pass and low-pass stages plus amplification); the trainable parameters are the corner frequencies of the high- and low-pass stages and the gain. Several such basis functions are summed to form one flexible edge. \textbf{b}, A small example PhyKAN, initialised with architecture [2, 3, 2, 1] and shown after pruning to [2, 1, 2, 1], trained to approximate the dimensionless Feynman function I.50.26 \cite{udrescu2020aifeynman} ($Y = f(X_0, X_1) =\mathrm{cos}(X_0)+X_1 \mathrm{cos}^2(X_0)$). Only the edges that survive training are drawn, each carrying its learned band-pass response (shown as red lines inside boxes); nodes (green circles) sum their incoming edges and apply a fixed sigmoid. This network is deliberately small so the mechanism is legible, and is illustrative rather than the network used for the hardware result in \textbf{c}. \textbf{c}, Hardware-realised output (left) of a separate, larger trained PhyKAN (architecture [2, 5, 5, 1]) for the same function, beside the true function (right), over the model's input domain. Colour scale is shared between the two maps. Hardware test MSE $\approx 1.76\times10^{-4}$.}
    \label{fig:schematic_overview}
\end{figure}
Fig.~\ref{fig:schematic_overview} shows that trainable nonlinear connections can be constructed from analogue electronic filters and transferred to hardware. The architecture places a learnable univariate function on each connection---the same arrangement as a multilayer perceptron (MLP), but with the scalar weights replaced by trainable nonlinearities, as motivated by the Kolmogorov--Arnold representation theorem\cite{liu2024kan}. Each edge is built from tuneable band-pass filter units whose inputs are encoded as signal frequencies (Fig.~\ref{fig:schematic_overview}a). Each node applies a fixed sigmoid function to bound inter-layer signals within the hardware operating range; the edge transfer functions provide the trainable nonlinearities. The filter corner frequencies and gains are the trainable parameters of the physical network (hereafter PhyKAN); the circuit, its analytical transfer-function model and the training procedure are given in Section~\ref{sec:Methodology}.

To make each edge more expressive, several filter units are summed in parallel and then sparsified during training, pruning units that contribute little (Section~\ref{sec:regularisation}; the regularised-MLP control is reported in Supplementary Note~\ref{sec:MLP_regularisation}). Fig.~\ref{fig:schematic_overview}b shows a PhyKAN initialised with architecture [2, 3, 2, 1] and trained to approximate a dimensionless Feynman function\cite{udrescu2020aifeynman}. Pruning reduces it to [2, 1, 2, 1], and the figure shows the pruned network, with nodes shown as green circles, and the response of each surviving edge carrying its learned band-pass response plotted as the red curves in black boxes.

The trained parameters are transferred directly to programmable analogue hardware, where the response of each edge in a network is measured using a fine-grained frequency sweep. Before deployment, the simulated parameters are mapped to the nearest hardware-achievable values---the corner frequencies through a look-up table of realisable filter settings, and the gains by rounding to the nearest achievable step---with a straight-through estimator maintaining a differentiable computational graph during training (Section~\ref{sec:s_t_e}). Fig.~\ref{fig:schematic_overview}c shows the function-fitting performance of the network output reconstructed from hardware-measured edge responses; the hardware output closely matches the ground-truth function, replicating the function with a mean-squared-error of 1.76$\times 10^{-4}$ calculated on a uniformly spaced grid of $[100\times100]$ samples over the range of the two input dimensions.
\subsection{Six-axis robotic kinematics}\label{sec:robotics}
\begin{figure}[!htbp]
    \centering
    \includegraphics[width=0.80\linewidth]{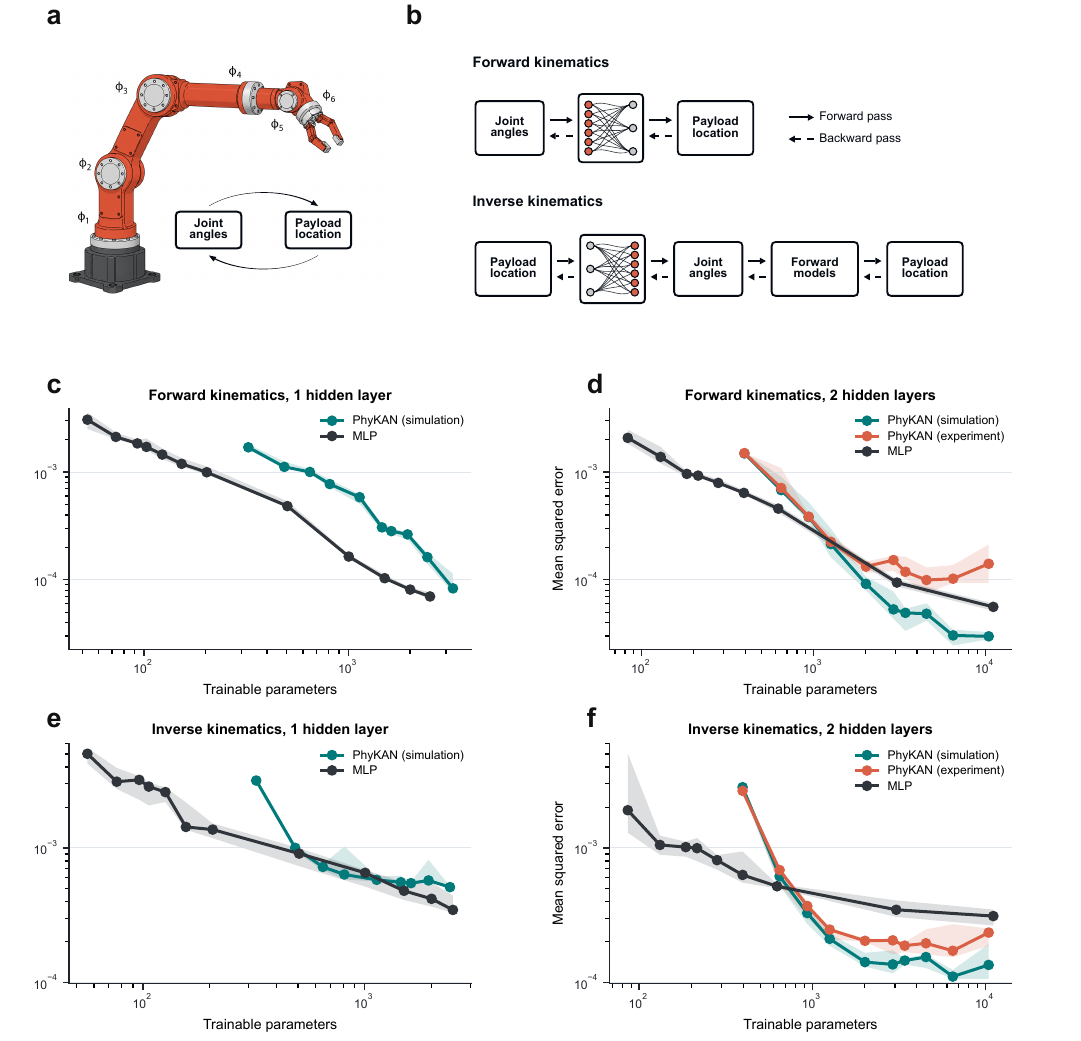}
    \caption{\textbf{Six-axis robot arm kinematics with PhyKANs.} \textbf{a}, Schematic of the joint configuration of the six-axis IRB120 commercial robotic manipulator modelled here. Rotational axes are shown in grey, where $\phi_i$ denotes the joint angle at each joint and black arrows show the plane of rotation at each joint. The payload (end-effector) location is determined by the centerpoint of the claw shown in the figure. \textbf{b}, Learning procedure for forward-kinematics modelling (upper) and inverse-kinematics position control (lower). In the forward-modelling problem, joint angles map uniquely to payload locations and the network approximates this map directly. In the inverse-control problem, the network predicts joint angles, which are passed through a frozen forward model---learned from the same noisy measurements and of the same architecture as the inverse network---to obtain an estimate of achieved location to enable optimisation; the test error is computed between the desired and achieved payload locations using the unseen, noiseless forward kinematic equations. \textbf{c--f}, Test mean-squared error versus trainable parameters for forward-kinematics modelling (\textbf{c}, \textbf{d}) and inverse-kinematics control (\textbf{e}, \textbf{f}), for one hidden layer (\textbf{c}, \textbf{e}) and two hidden layers (\textbf{d}, \textbf{f}): PhyKAN (simulation; teal), PhyKAN (experiment; orange, where shown), and multilayer perceptrons (black). Experimental transfer is shown only where the simulated PhyKANs offer a parameter advantage. Points show medians and shaded regions show interquartile ranges across independently trained models: 10 models per simulated PhyKAN and MLP point, and five transferred models per experimental PhyKAN point.}
    \label{fig:Robotics}
\end{figure}
Smooth edge nonlinearities can be composed to form multivariate functions, making them particularly well suited for representing the kinematics of rigid bodies with high parameter efficiency. In this study, we model the forward kinematics and learn an inverse-kinematics position controller for a robotic manipulator using filter-based PhyKANs. Fig.~\ref{fig:Robotics}a shows a schematic diagram of the IRB120 commercial robotic manipulator manufactured by ABB, which has been previously used for neural-network-based approximation of kinematics\cite{vsegota2021utilization}, and is used here to benchmark the PhyKANs.

The manipulator is controlled by six revolute joints that move the end-effector in three-dimensional space. By perturbing each joint angle independently with noise sampled uniformly over $\pm0.1$\,rad in the forward kinematics equations, we emulate the imperfect actuation of physical motors. The joint angles and arm lengths uniquely define the end-effector location, so we built a forward-kinematics dataset by computing end-effector locations for joint angles sampled randomly across the ranges given in Section~\ref{sec:kinematics_data}. The Denavit--Hartenberg parameters used to determine these end-effector locations via a spatial kinematic chain are outlined in the same section. During training, the networks see only the ideal joint angles as inputs, while receiving end-effector locations that reflect imperfect actuation. Throughout, reported kinematics errors are evaluated using the noiseless equations, isolating systematic prediction or control error from irreducible trial-to-trial actuator variability. For the forward task, this measures how well each network recovers the nominal map from actuator-corrupted training targets. In the noiseless comparison, unconstrained PhyKANs retain an advantage over MLPs at larger network sizes; hardware discretisation reduces this advantage in forward kinematics, while it is retained in the reported inverse-kinematics comparison (Supplementary Fig. \ref{fig:noiseless_robotics}).

We formulate inverse kinematics as a model-based feedforward position-control problem: given a target end-effector location, the network predicts the joint-angle commands required to reach it. Multiple joint-angle solutions exist for a given end-effector location, so simply reversing the inputs and targets of the forward problem is ambiguous. The kinematic equations, together with the actuation noise, define the ground-truth system, which the network observes only through sampled measurements. To train the inverse model, we first learn a forward model from these measurements; this forward model is then frozen and used to close the training loop. Specifically, the inverse network predicts joint angles, which are passed through the frozen forward model to obtain an estimated end-effector location, and the loss is computed between the target location and this estimate, as shown in  Fig.~\ref{fig:Robotics}b. Each inverse network is trained through a frozen forward model of its own architecture, both learned from the same measurements. For inverse control, the same evaluation applies the noiseless kinematic equations to the predicted joint commands and measures the resulting end-effector error.

Fig.~\ref{fig:Robotics}c,d show the achieved mean-squared error between model-predicted end-effector location and true location given by the kinematic equations as a function of trainable parameters for one and two hidden layers respectively. Trainable-parameter counts are computed from the unpruned architecture as initialised, so that they account for all parameters available to the network during learning rather than only those surviving pruning. Teal points show the PhyKAN models in simulation, compared to standard multilayer perceptrons (MLPs, black) with ReLU activation functions as a baseline for conventional neural networks. Although each edge carries 18 trainable parameters, the PhyKANs need substantially fewer hidden-layer nodes in simulation, giving greater parameter efficiency with multiple hidden layers and an advantage in the number of connections throughout (see Fig. \ref{fig:robotics_vs_edgecount}). The greater accuracy offered by deeper versus wider networks corroborates findings reported for other physical KANs\cite{taglietti2026}. However, due to experimental mismatch, the experimental PhyKANs underperform the MLPs for the forward kinematics problem at the largest network sizes, by approximately a factor of two, while the most accurate simulated networks reach mean-squared errors $<1\times10^{-4}$.

Fig.~\ref{fig:Robotics}e,f compare positioning errors for the inverse-kinematics control task between filter-based PhyKANs and standard MLPs for networks with one and two hidden layers respectively. For networks with one hidden layer, the MLPs outperform the PhyKANs at the smallest and largest network sizes, with the two closely matched in between. However, in the two-hidden-layer networks, beyond $\approx1,000$ parameters (5 nodes per layer, 18 parameters per edge), the PhyKANs achieve lower error than the MLPs, and this advantage persists after hardware transfer. This pattern is consistent with the hypothesis that smooth nonlinear edge functions favour interpolation through actuator-corrupted observations.

In the two-hidden-layer kinematics tasks, the simulated PhyKAN error decreases with parameter count, with diminishing returns at larger model sizes (Fig.~\ref{fig:Robotics}d,f). A descriptive analysis of the intrinsic dimensionality of the hidden representations of both architectures, using the separability-based intrinsic dimensionality metric of Sutton et al.~\cite{Sutton_intrindim}, is reported in Supplementary Note~\ref{sec:dimensionality}.
\subsection{Continuous-action control}\label{sec:RL}
As well as approximating known input--output relationships, PhyKANs can model decision-making policies learned from environmental feedback in a reinforcement-learning paradigm. Here, we use an actor--critic algorithm to learn policies which maximise a given reward in an environment with a continuous action space. In this paradigm, a `critic' network estimates total discounted future rewards from a given state--action pair, while an `actor' network learns a policy for action selection given a current state, maximising estimates of reward generated by the critic. Full implementation details can be found in Section~\ref{sec:AC_training}.

\begin{figure}[!htbp]
    \centering
    \includegraphics[width=\linewidth]{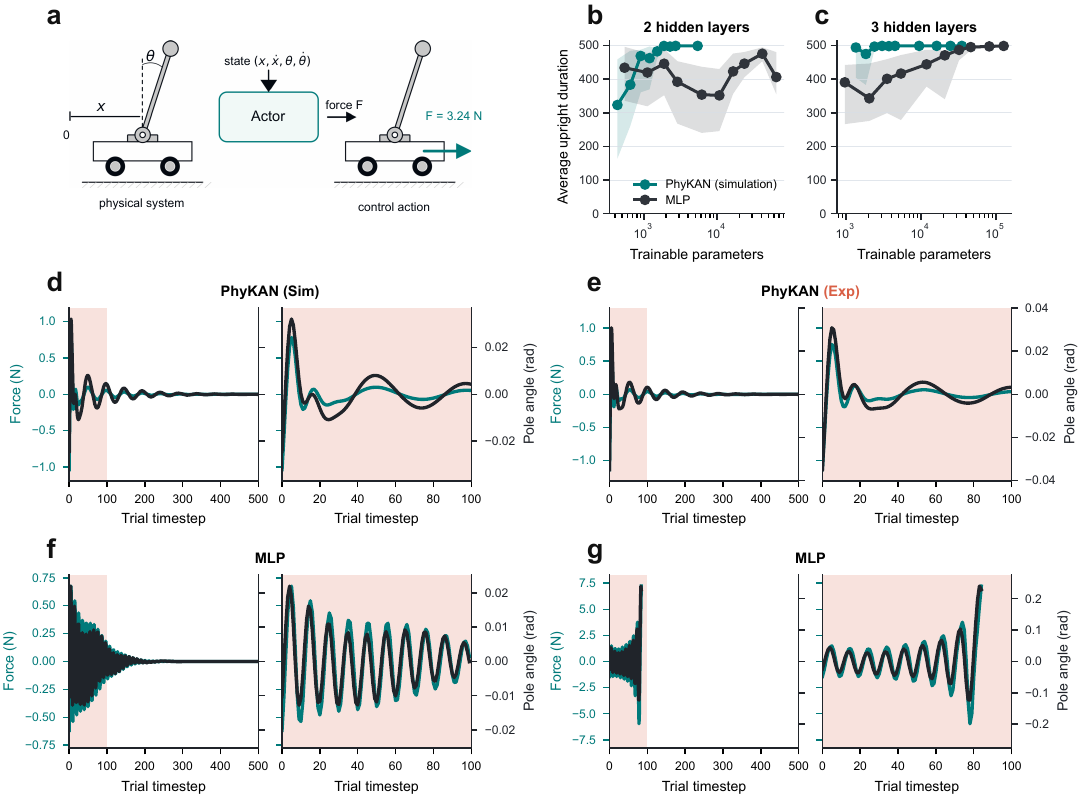}
    \caption{\textbf{Continuous-action CartPole control with PhyKANs.} \textbf{a}, Schematic of the CartPole task. The actor provides a continuous output corresponding to a vector force applied to the cart, with the goal of keeping the cart position, $x$, and pole angle, $\theta$, close to zero. \textbf{b}, \textbf{c}, Average duration for which the agent maintains the pole upright versus trainable parameters, for PhyKAN (simulation; teal) and MLPs (black), with two hidden layers (\textbf{b}) and three hidden layers (\textbf{c}). Performance for each trained model is averaged over 100 initial conditions, with the right-sided panels showing a zoomed view of the shaded regions on left-sided panels. Points show medians and shaded regions show interquartile ranges across 10 independently trained PhyKAN models and 20 independently trained MLP models; more MLP repetitions were run because of their greater across-model variance. \textbf{d--g}, Example runs of the CartPole environment, showing force applied (teal) and pole angle (black) over the first 100 timesteps. \textbf{d}, \textbf{e}, Typical runs for PhyKAN (simulation) and PhyKAN (experiment), respectively, both of which solve the task. \textbf{f}, \textbf{g}, Characteristic solution and failure cases, respectively, for an MLP network that maintains the pole upright for $\approx$75\% of initial conditions.}
    \label{fig:CartPole}
\end{figure}

As a demonstration, we perform a classical reinforcement-learning control task, CartPole, modified to use continuous-valued forces instead of a binary choice between $\pm10$ N. Here, the environment models a pole attached to a cart via a revolute joint, shown schematically in Fig.~\ref{fig:CartPole}a. The actor network uses observations of the position and velocity of the cart ($x$ and $\dot{x}$), as well as the angle and angular velocity of the pole ($\theta$ and $\dot{\theta}$), to provide a continuous force in the positive (right) or negative (left) direction that maximises the expected total future reward estimated via the critic network. In this task, the agent receives increased rewards for maintaining the pole as close to upright as possible, up to a maximum trial length of 500 timesteps. The critic network is updated using temporal-difference learning to align its predictions with the rewards actually received. We consider the task solved when the agent consistently maintains the pole upright for 500 timesteps.

Fig.~\ref{fig:CartPole}b,c compare performance on the continuous-action CartPole task between PhyKANs (simulation; teal) and conventional MLPs (black) as a function of trainable parameters for two and three hidden layer networks respectively. Performance is measured by training the networks for 1000 episodes, and evaluating each trained agent over 100 randomly sampled initial conditions. Plotted data points show the median of this performance measure across 10 independently trained PhyKAN models and 20 independently trained MLP models, while the shaded regions show the corresponding interquartile ranges. In both cases, the PhyKANs solve the task with fewer trainable parameters than MLPs, demonstrating the architecture's suitability for continuous-action reinforcement learning.

Fig.~\ref{fig:CartPole}d--g compare the typical behaviours learned by PhyKAN and MLP networks with $\approx2,000$ trainable parameters and two hidden layers, where the PhyKAN has solved the task optimally, while the MLP has discovered a suboptimal solution in which the agent maintains the pole upright for the full 500 timesteps for around 75\% of the sampled initial conditions. Fig.~\ref{fig:CartPole}d shows that the simulated PhyKAN quickly establishes a stable pole angle from initial conditions, and maintains the pole upright with negligible deviation. When transferred to experimental hardware (Fig.~\ref{fig:CartPole}e), small mismatches in experimentally realised edges lead to different trajectories due to the different actions taken, though with feedback from the environment, the agent can still stabilise the pole. On the other hand, the MLP (Fig.~\ref{fig:CartPole}f,g) exhibits higher-frequency oscillatory behaviour, as the direction of the applied force reverses rapidly in response to changing states, pointing towards an unstable solution. Depending on the initial conditions, the pole is either stabilised (Fig.~\ref{fig:CartPole}f), or the oscillations grow until the trial fails (Fig.~\ref{fig:CartPole}g).
\FloatBarrier
\begin{figure}[!htbp]
    \centering
    \includegraphics[width=\linewidth]{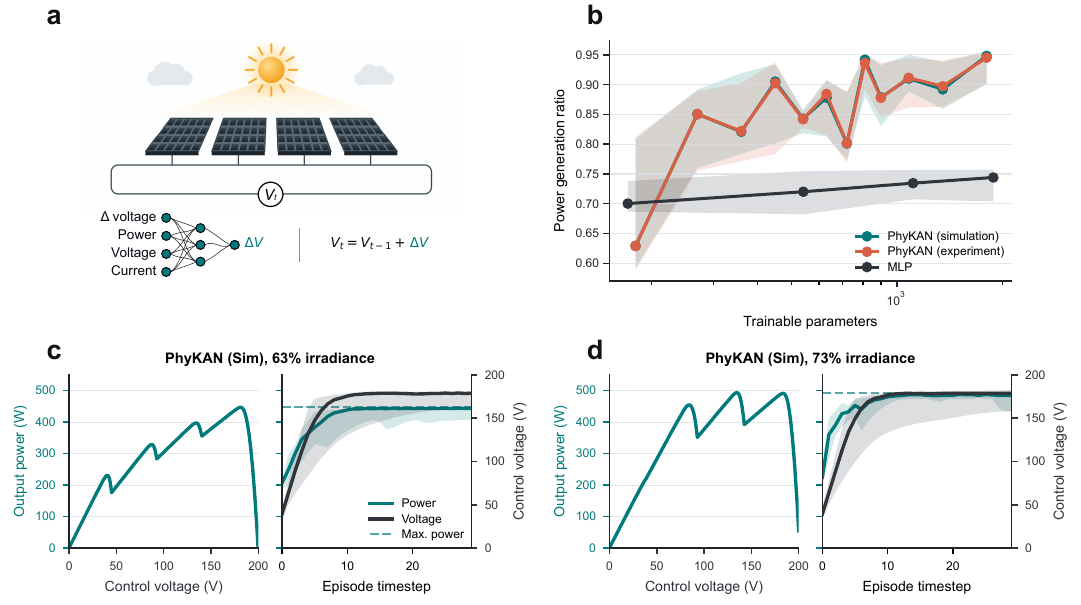}
    \caption{\textbf{Maximum power point tracking for a photovoltaic array under partial shading conditions (MPPT).} \textbf{a}, Schematic of the MPPT task. A photovoltaic array operates under variable irradiance. A state vector of current, voltage, power and previous change in voltage is passed to the controller network, which outputs a change in control voltage applied to a voltage-booster circuit, changing the generated power and current across the arrays. \textbf{b}, Power generation ratio versus trainable parameters for PhyKAN (simulation; teal), the MLP (black), and PhyKAN (experiment; orange). PhyKAN (experiment) denotes a controller reconstructed from experimentally measured edge responses and evaluated in the modelled photovoltaic environment. Power generation ratio is the power output of the controlled system divided by the maximum available power for a given irradiance condition, calculated from physical models. Performance for each trained model is averaged over 100 irradiance conditions; points show medians and shaded regions show interquartile ranges across 10 independently trained models. \textbf{c}, \textbf{d}, Example PhyKAN (simulation) controller responses to two partial-shading conditions of different complexity (four local power maxima in \textbf{c}, three in \textbf{d}). Left panels: modelled power (black) versus control-voltage response of the array, with the maximum available power marked by a dashed line. Right panels: controller trajectory over the first 30 timesteps, showing output power (teal) and control voltage (black); lines show medians and shaded regions show interquartile ranges across 10 independently trained models. Each condition is annotated with its mean module irradiance relative to full sun.}
    \label{fig:PV_Control}
\end{figure}
We next consider photovoltaic maximum-power-point tracking, a power-electronics control problem in which compact, low-power inference is directly relevant. The objective in this task is to change the control voltage held across a photovoltaic array to maximise the output power of the array while remaining agnostic to the irradiance incident on each cell. Input states consist of the applied control voltage, the output current of the array, the change in voltage compared to the previous timestep, and the generated power. Details of the simulation that generates the PV curves used as state data for the environment can be found in Section~\ref{sec:PV_data}.

Fig.~\ref{fig:PV_Control}b compares the generated power ratio between PhyKAN (simulation; teal), MLPs (black), and PhyKAN (experiment; orange) as a function of trainable parameters. Here, power generation ratio is defined as the ratio between the power being generated and the maximum available power for the current irradiance condition, determined via simulation. The data presented reflect average performance across 100 randomly sampled irradiance conditions, each with randomly generated initial conditions and 50 timesteps during which the agent can change the control voltage. The plotted data points show the median performance across 10 independently trained models, and the shaded regions show the corresponding interquartile ranges. Except for the smallest network, with only 2 hidden nodes, the PhyKANs maintain a performance advantage over the MLP baselines across the tested parameter range, with the gap persisting at the largest network sizes evaluated. The persistent gap reflects a greater ability to find the maximum power point quickly and reliably from environmental feedback.

Fig.~\ref{fig:PV_Control}c,d show example trials for the PhyKAN controllers in simulation under two different partial shading conditions: the modelled power versus control-voltage response for each condition (left panels), and the corresponding controller trajectory over the first 30 timesteps (right panels). In both cases the curves contain local maxima, which the agent avoids while finding the maximum power point, a known challenge for simple control methods such as perturb-and-observe algorithms \cite{mohapatra2017review}.

\subsection{Conditions for the trainable parameter advantage in PhyKANs}\label{sec:task_dependence}
The kinematics and continuous-action control tasks (Sections~\ref{sec:robotics} and~\ref{sec:RL}) all require the network to represent smooth, continuously valued mappings, and in each the supervision is imperfect: actuator-corrupted observations in the kinematics tasks, and bootstrapped value estimates or environmental feedback in the reinforcement-learning control tasks. In a separate comparison, with noiseless kinematic targets, the unconstrained networks retain an advantage at larger network sizes, while the hardware-discretised networks retain it in inverse kinematics but not in forward kinematics (Supplementary Fig.~\ref{fig:noiseless_robotics}).

On two decision-based tasks---Fashion-MNIST classification and the standard CartPole task with binary ($\pm10$ N) actions---a consistent advantage does not appear (Supplementary Figs.~\ref{fig:FMNIST} and~\ref{fig:binary_CP}). On Fashion-MNIST the per-parameter advantage over MLPs is largely absent, surviving only at the smallest network sizes, and the number of filters per edge has little effect on accuracy. The remaining advantage is in network connectivity: PhyKANs achieve a given accuracy with fewer edges (connections) than MLPs. On binary CartPole the two architectures are closely matched---with one hidden layer they reach maximum performance at similar parameter counts (the MLP leading only at the smallest sizes), and with two hidden layers performance is almost identical. Because greedy selection between two discrete actions is a winner-take-all decision on the predicted Q-values, this task has a decision-boundary component similar to binary classification.

Two mechanisms are consistent with these results, and both may be acting. One possible mechanism concerns the interaction between hardware discretisation and the supervision used during training. On the FPAA, filter corner frequencies and gains can take only values from a finite hardware grid. Hardware-aware training therefore uses these quantised values in the forward pass while propagating gradients through continuous latent parameters using a straight-through estimator. Under exact supervision, repeatedly using the same quantised value can make the surrogate updates depend systematically on the parameter's position relative to the hardware grid, limiting the precision attainable during optimisation. Variability in the training signal may disrupt this systematic relationship and assist movement between neighbouring hardware values. This would play a role analogous, although not identical, to stochastic rounding in low-precision training\cite{gupta2015limited,courbariaux2015binaryconnect} and to dither that decorrelates quantisation error from the signal being quantised\cite{lipshitz1992dither}. Our results are consistent with this interpretation: across the continuous tasks, hardware-discretised networks retain their advantage when training involves actuator-corrupted observations, bootstrapped value estimates or a learned intermediate model, whereas the advantage diminishes with network size in noiseless forward kinematics, where training is performed directly against exact targets (see Supplementary Figure \ref{fig:noiseless_robotics}).

The second is the form of the target. A multilayer perceptron with piecewise-linear units partitions the input space with hyperplanes\cite{montufar2014linear}, so a task that reduces to separating regions is directly matched to what the architecture builds. A bank of band-pass filters has no such primitive: to approximate a sharp boundary, many smooth functions must be combined, which requires a larger number of filters. The advantage of trainable nonlinear edges therefore requires a continuously valued target. Where the output is instead a discrete choice---Fashion-MNIST, and binary CartPole, where greedy action selection is a comparison of two predicted Q-values---the multilayer perceptron is favoured, and no consistent parameter-efficiency advantage is observed.

Neither mechanism alone accounts for all of the results. The first does not explain binary CartPole, where the supervision is bootstrapped yet no advantage appears; the second does not explain the noiseless kinematics, where the target is smooth yet the discretised networks lose ground. Together they provide a consistent interpretation of the remaining results, with the exception of the one-hidden-layer networks, where the parameter-efficiency advantage requires sufficient depth, although the advantage in the number of connections is retained (Fig.~\ref{fig:Robotics}c,e; Supplementary Fig.~\ref{fig:robotics_vs_edgecount}).
\FloatBarrier
\subsection{Memristor-based PhyKANs}\label{sec:memristors}
\begin{figure}[htbp!]
    \centering
    \includegraphics[width=\linewidth]{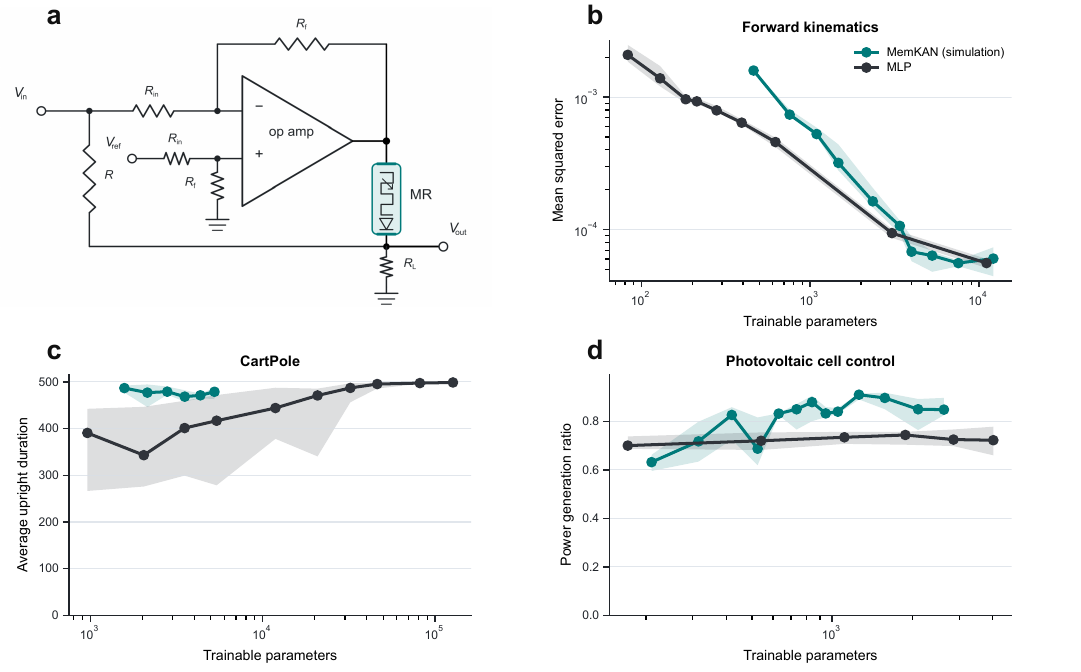}
    \caption{\textbf{Memristor-based PhyKANs.} \textbf{a}, Circuit topology of the underlying nonlinear element, showing the input and reference voltages, operational amplifier, memristor (MR), and output load. \textbf{b--d}, Performance comparison between MLPs (black) and MemKAN (simulation; teal) versus trainable parameter count, for the six-axis forward kinematics task (\textbf{b}), the CartPole task (\textbf{c}), and the photovoltaic-cell control task (\textbf{d}). Both models have two hidden layers in \textbf{b} and three hidden layers in \textbf{c}; in \textbf{d}, the MemKAN has one hidden layer and the MLP has two because the one-hidden-layer MLP did not achieve competitive performance on this task. Points show medians and shaded regions show interquartile ranges across independently trained models. Panels \textbf{b} and \textbf{d} use 10 models per point for both architectures; panel \textbf{c} uses 10 MemKAN models and 20 MLP models per point.}
    \label{fig:MemKAN}
\end{figure}

To test whether the same trainable-nonlinear-edge principle extends beyond analogue filters, we replaced the analogue filters that provide the tuneable transfer functions with simulated memristor circuits (MemKAN). The three continuous tasks performed with the analogue-filter networks were then repeated with memristor-based networks.

The memristors modelled here are two-terminal devices whose current--voltage relationship depends on an internal state variable, giving a programmable and strongly nonlinear response \cite{Chua1971,Strukov2008}; the device physics and the model used to capture it are described in Section~\ref{sec:memdiode_model}.

For the memristor-based PhyKAN, both inputs and outputs of each nonlinear synapse are encoded in pulse amplitude, not in the frequency domain. To obtain transfer functions with non-monotonic transfer characteristics, and hence turning points, we use the programmable nonlinear current--voltage behaviour of a memristor operated in its high-resistance state and embedded in a reconfigurable active network that can realise tuneable nonlinear functions. This is achieved through a modified subtractor circuit configuration in which an output-sensing resistor is additionally driven by a current proportional to $(V_{\mathrm{in}} - V_{\mathrm{out}})/R$, introducing a feedback mechanism from the voltage across the memristor that selectively enhances intermediate input amplitudes while attenuating both low and high extremes. Fig.~\ref{fig:MemKAN}a shows the underlying components behind the memristor edges.

The electrical transport properties of the circuit were determined by a dynamic memdiode model\cite{Aguirre2022} (see Methods, Section~\ref{sec:memdiode_model}) to resolve the state of the memristive element, coupled with a circuit solver for the interplay between the memristor and the operational amplifier. The model requires an iterative nonlinear least-squares step to find the I/V response. We therefore used a surrogate model, which provides a non-recursive differentiable map from input voltages to output voltages given the control parameters. This surrogate took the form of a standard MLP, with further details in Section~\ref{sec:memristor_MLP}.

Fig.~\ref{fig:MemKAN}b shows performance on the forward kinematics task of Fig.~\ref{fig:Robotics}, while panels c and d show the CartPole and PV-control tasks respectively. On the forward kinematics task the MLP holds the advantage up to approximately $3\times10^{3}$ trainable parameters, beyond which the MemKAN achieves lower error. As with the filter-based networks, the MemKANs show improved parameter efficiency over the MLP controls in the CartPole and PV-control tasks, with performance comparable to the filter-based PhyKANs. The similar results from two different nonlinear edge models support the hypothesis that placing trainable nonlinearities on network edges contributes to the improved efficiency observed here and in other studies \cite{wen2026computing, escudero2026physical,taglietti2026}.

\section{Discussion}\label{sec:discussion}

Across the tasks tested here, the clearest common condition for parameter-efficiency gains is a smooth, continuously valued mapping. On the Fashion-MNIST and binary-action CartPole benchmarks (Supplementary Figs.~\ref{fig:FMNIST} and \ref{fig:binary_CP}), which contain decision-boundary or winner-take-all components, the architecture performs comparably to a multilayer perceptron. On the smooth, continuously valued tasks studied here, the architecture concentrates computation into expressive physical edges, reducing the number of nodes, active connections and trainable parameters needed to reach a given performance in these tasks; when mapped to hardware, this means fewer analogue edge circuits to route, program and calibrate. This pattern is observed across robotic kinematics learned from actuator-corrupted observations, continuous-action reinforcement learning with bootstrapped targets, and photovoltaic maximum-power-point tracking learned from environmental feedback, with the strongest effect in the last of these, where the nonlinear-edge networks outperform multilayer perceptrons at all but the smallest network size.

The hardware results indicate that this architectural advantage can survive transfer from model to device for the closed-loop control tasks demonstrated here, with diminishing advantage for the regression problems. Training is performed through analytical transfer-function models of the analogue filters, with hardware-aware discretisation of trainable parameters before deployment. Pooling the edges of the networks transferred here gives approximately 35,000 experimentally realised connections, of which 90\% match their simulated targets to within a mean-squared error of $7.92 \times 10^{-5}$. The remaining error is mainly due to parasitic capacitance and resistance that are not captured by the analytical model. These results use direct transfer only: after mapping the simulated parameters to hardware-achievable values, we did not carry out hardware-in-the-loop hill climbing or reinforcement-learning fine-tuning. Such post-transfer calibration could further reduce residual mismatch, particularly for larger networks where transfer errors may compound. This mismatch could be further reduced by favouring, during training, parameter values for which small deviations leave the edge response essentially unchanged, as in sharpness-aware training\cite{xu2026physical}. Closed-loop tasks provide an additional tolerance mechanism, because the controller can compensate for small deviations through subsequent interactions with the environment. Scaling to larger networks will require more complete modelling of parasitics or post-transfer fine-tuning, but the present results show that direct model-to-hardware transfer is already accurate enough for the closed-loop control tasks demonstrated here, where the controller compensates for residual deviations. Open-loop function approximation is more sensitive: the transferred forward-kinematics models do not retain their advantage over multilayer perceptrons (Fig.~\ref{fig:Robotics}d).

The same design principle extends beyond the analogue-filter platform. In simulated memristor-based networks, the trainable nonlinearity is provided by the steady-state current--voltage characteristic of the device rather than by a frequency-domain filter response. Across the three continuous tasks evaluated with this model, the MemKAN shows parameter-efficiency trends similar to those of the filter-based PhyKAN, including a persistent advantage over multilayer perceptrons in photovoltaic tracking. These results support the cross-substrate applicability of trainable nonlinear connections on the continuous tasks examined here, as demonstrated for both analogue-filter and memristor-based networks. For memristive computing, this is a substantial shift in emphasis. Memristors are commonly engineered to behave as programmable linear conductances in crossbar arrays, suppressing the nonlinear current--voltage response and often pushing operation into higher-current regimes\cite{joksas2022nonideality}. Here, that nonlinear response becomes the computational element itself, allowing operation in the natural low-current analogue regime where power dissipation is reduced and endurance can be improved. The reduction in the number of connections may translate into a simpler physical layout where the nonlinear edge can be integrated as a compact cell, and a two-terminal device embedded in a small active circuit is closer to this than a bank of reconfigurable filters.

The experimental platform used here was chosen for reconfigurability and measurement access, not maximum efficiency. On dedicated hardware, the reduced network size enabled by nonlinear physical edges is projected to translate into lower power consumption. Because a nonlinear edge is physically more complex than a scalar weight, the hardware benefit cannot be inferred from parameter count alone. We therefore estimate power from the number of required edge circuits and a component-level CMOS design of those circuits (Section~\ref{sec:power_cons}). Based on this CMOS-compatible transconductor implementation, the photovoltaic control networks are projected to consume ${\sim}35\;\mathrm{\mu W}$, around two orders of magnitude below the $0.33$--$22\;\mathrm{mW}$ envelope of microcontrollers used for comparable edge-computing tasks. This estimate combines the intrinsic efficiency of analogue operation with the architectural reduction in required components.

Overall, this work points to an alternative route for physical neural networks. Rather than forcing analogue devices to approximate software-style scalar weights, their native nonlinear responses can be made trainable and used directly as the computational primitives of the network. For the smooth, continuous inference and control tasks studied here, this produces compact networks that transfer to analogue hardware and can be mapped onto distinct physical substrates, suggesting a promising strategy for low-power physical machine learning.

\section*{Acknowledgements}
This work was primarily supported by the UK Neuromorphic Computing Hardware Semiconductor IKC (Neuroware; UKRI2784), funded by EPSRC and Innovate UK. I.T.V., A.J.K., A.Me. and E.V. acknowledge support from Neuroware. Additional support was provided by the EPSRC MARCH project: I.T.V., L.M., M.O.A.E., T.J.H. and E.V. acknowledge support from EP/V006339/1, and M.T. and S.S. acknowledge support from EP/V006029/1. E.V. and L.M. also acknowledge support from the CHIST-ERA project Causal eXplanations in Reinforcement Learning (CausalXRL; CHIST-ERA-19-XAI-002), funded by EPSRC under grant EP/V055720/1.

\section*{Author Contributions}
E.V. conceived and initiated the physical Kolmogorov--Arnold network approach investigated in this study. I.T.V. proposed and implemented the analogue-filter implementation. I.T.V. performed the simulations, experiments and task implementations with input from E.V. E.V. contributed to software development and validation of the computational results. I.T.V. and E.V. analysed the results. E.V. proposed photovoltaic maximum-power-point tracking as an application, and I.T.V. implemented the photovoltaic-control task. I.T.V., T.J.H., J.C.G. and E.V. refined the study through discussion. I.T.V., F.A., A.J.K., A.Me. and E.V. developed the memristor-based realisation. F.A. developed the memdiode model and circuit topology with input from A.Me. I.T.V., C.S. and T.J.H. designed and implemented the measurement infrastructure. I.T.V. and M.O.A.E. developed the FPAA programming code. I.T.V., A.Mc., M.T. and T.J.H. developed the FPAA deployment. I.T.V. and E.V. conducted the intrinsic-dimensionality analysis. O.J.S., L.M., F.R. and E.V. performed preliminary evaluations of the intrinsic-dimensionality metric on other network architectures, with input from I.Y.T. E.V. and I.Y.T. developed the universality argument with input from O.J.S. I.T.V. and E.V. drafted the manuscript. F.A. drafted the memristor sections. T.J.H., J.C.G., M.T., O.J.S., M.O.A.E., L.M. and S.S. provided feedback on the manuscript. E.V. directed the research.

\section{Methods}\label{sec:Methodology}
\subsection{Definition of a Physical KAN-type Edge}
A PhyKAN is built from trainable nonlinear edges. Consider an edge from node $i$ in layer $\ell$ to node $h$ in layer $\ell+1$. This edge receives the scalar activation $a_i^{(\ell)}$ and returns a learned univariate response formed by summing the outputs of a bank of analogue band-pass filters:
\begin{equation}
\Phi_{i,h}\bigl(a_i^{(\ell)}\bigr)
= \sum_{k=1}^{K_{i,h}} G_{i,h,k}\,
\rho\!\left(f_{\mathrm{enc}}\bigl(a_i^{(\ell)}\bigr);\vartheta_{i,h,k}\right),
\end{equation}
where $K_{i,h}$ is the number of filters on the edge, $G_{i,h,k}$ is the learned gain of filter $k$, $\vartheta_{i,h,k}$ denotes its learned corner-frequency parameters, and $\rho(\nu;\vartheta)\coloneqq |H_{\mathrm{BP}}(\nu;\vartheta)|$ is the corresponding band-pass magnitude response. The complete network is obtained by summing these edge responses at each receiving node and applying the fixed sigmoidal conditioning described below.

\subsection{PhyKAN: Implementation Details}
In our implementation, each single-filter response $\rho(\nu;\vartheta)$ is realised by a cascade of first-order high-pass and low-pass stages in steady state. The following subsections describe the constituent components used to implement a filter-based PhyKAN.

The illustrative network in Fig.~\ref{fig:schematic_overview}b uses an ideal continuous-filter parameterisation, without hardware quantisation or frequency warping. The hardware result in Fig.~\ref{fig:schematic_overview}c and the other hardware-transfer results use the hardware-aware parameterisation described below.
\subsubsection{Input-to-Frequency Encoding}
To ensure inputs remain within the operating range of hardware, raw inputs to the network, $\hat{x}$, are constrained via the scaled sigmoid $\sigma_s$ defined below if their ranges exceed $[0,1]$:
\begin{equation}
x = \sigma_s(\hat{x}) = \frac{1}{1+e^{-\hat{x}/s_\sigma}}.
\end{equation}
Inputs that are already bounded to $[0,1]$ are passed directly to the encoding map without this step; this is the case for the robot kinematics tasks, whereas the reinforcement-learning tasks, whose state variables are unbounded, use the sigmoid.
For a constrained input $x \in [0, 1]$, we define the encoding map
\begin{equation}
f_{\mathrm{enc}}(x) = 10^{\alpha_{\text{in}} + \beta_{\text{in}} x},
\end{equation}
where $\alpha_{\text{in}} = 3.65$ is the log-frequency offset and $\beta_{\text{in}} = 1.5$ is the log-frequency slope. This maps roughly to $10^{\alpha_{\text{in}}} = 10^{3.65} \approx 4470 \,\text{Hz}$ (for $x = 0$) up to $10^{\alpha_{\text{in}} + \beta_{\text{in}}} = 10^{5.15} \approx 141 \,\text{kHz}$ (for $x = 1$).
\subsubsection{Learnable Filter Parameters}
Each filter unit on edge $(i,h)$ with index $k$ has three trainable parameters stored as
\begin{equation}
\theta_{i,h,k} = \bigl(g_{i,h,k},\, p_{i,h,k}^{\text{low}},\, p_{i,h,k}^{\text{high}}\bigr).
\end{equation}

Here $g_{i,h,k}, p_{i,h,k}^{\text{low}}, p_{i,h,k}^{\text{high}} \in \mathbb{R}$ are unconstrained (raw) parameters for the gain and the low-pass and high-pass corner frequencies of filter $k$ on edge $(i,h)$. The corresponding physical gain $G_{i,h,k}$ and corner frequencies $\nu^{\text{low}}_{i,h,k}$ and $\nu^{\text{high}}_{i,h,k}$ are defined below. These raw parameters are bounded to values achievable on hardware via a scaled sigmoid
\begin{equation}
\sigma_s(z) = \frac{1}{1 + e^{-z / s_{\sigma}}},
\end{equation}
where $s_{\sigma} = 0.5$ sets the slope of the nonlinearity.
\paragraph{Gain.}
The effective gain applied by filter $k$ on edge $(i,h)$ is
\begin{equation}
G_{i,h,k} = G_{\max}\left(\sigma_s(g_{i,h,k}) - \tfrac{1}{2}\right),
\end{equation}
where $G_{\max} = 3$ fixes the overall gain scale, so that $G_{i,h,k} \in [-1.5, 1.5]$ for the parameterisation used here.

\paragraph{Corner frequencies.}
The low-pass and high-pass corner frequencies for filter $k$ on edge $(i,h)$, corresponding to the raw parameters $p_{i,h,k}^{\text{low}}$ and $p_{i,h,k}^{\text{high}}$ from $\theta_{i,h,k}$, are defined as
\begin{equation}
\nu^{\text{low}}_{i,h,k} = 10^{\alpha_{\text{c}} + \beta_{\text{c}} \, \sigma_s(p_{i,h,k}^{\text{low}})},
 \qquad
\nu^{\text{high}}_{i,h,k} = 10^{\alpha_{\text{c}} + \beta_{\text{c}} \, \sigma_s(p_{i,h,k}^{\text{high}})},
\end{equation}
where $\alpha_{\text{c}} = 3.65$ and $\beta_{\text{c}} = 1.9$ determine the range of available corner frequencies. For these values, the corner frequencies lie in
\begin{equation}
10^{\alpha_{\text{c}}} \approx 4.5 \,\text{kHz}
\quad \text{to} \quad
10^{\alpha_{\text{c}} + \beta_{\text{c}}} \approx 355 \,\text{kHz}.
\end{equation}
These bounds slightly exceed the input encoding range, allowing filters to place their effective passbands both within and near the edges of the driven frequencies.

\subsubsection{Band-pass Filter Transfer Function}\label{sec:transfunc}

The PhyKAN uses a cascade of first-order high-pass and low-pass filters, described by the following equations, which give the output-to-input amplitude ratio at steady state as a function of input frequency $\nu$.

\paragraph{High-pass filter.}
\begin{equation}
\label{eq:highpass}
H_{\text{HP}}(\nu) = \frac{j \omega_a \tau_H}{1 + j \omega_a \tau_H},
\qquad
\tau_H = R C_{\text{high}} = \frac{1}{2\pi \nu^{\text{high}}},
\end{equation}
where $\omega_a = 2\pi \nu$.

\paragraph{Low-pass filter.}
\begin{equation}
\label{eq:lowpass}
H_{\text{LP}}(\nu) = \frac{1}{1 + j \omega_a \tau_L},
\qquad
\tau_L = R C_{\text{low}} = \frac{1}{2\pi \nu^{\text{low}}}.
\end{equation}

\paragraph{Combined band-pass response.}
The combined (magnitude) response for a single filter is
\begin{equation}
H(\nu; \theta_{i,h,k}) = G_{i,h,k} \cdot \left| \frac{j \omega_a \tau_H}{1 + j \omega_a \tau_H}
\cdot \frac{1}{1 + j \omega_a \tau_L} \right|.
\end{equation}
Substituting $\tau_L = 1 / (2\pi \nu_L), \space \tau_H=1/(2\pi\nu_H)$ gives
\begin{equation}
H(\nu; \theta_{i,h,k}) =
G_{i,h,k} \cdot \left|
\frac{j \, (\nu / \nu^{\text{high}}_{i,h,k})}{1 + j \, (\nu / \nu^{\text{high}}_{i,h,k})}
\cdot
\frac{1}{1 + j \, (\nu / \nu^{\text{low}}_{i,h,k})}
\right|.
\end{equation}
In particular, the band-pass magnitude $H_{\mathrm{BP}}$ appearing in the definition of $\rho$ above is given by
\begin{equation}
H_{\mathrm{BP}}\!\left(\nu; \nu^{\text{low}}_{i,h,k}, \nu^{\text{high}}_{i,h,k}\right)
=
\left|
\frac{j \, (\nu / \nu^{\text{high}}_{i,h,k})}{1 + j \, (\nu / \nu^{\text{high}}_{i,h,k})}
\cdot
\frac{1}{1 + j \, (\nu / \nu^{\text{low}}_{i,h,k})}
\right|.
\end{equation}
In the network, the scalar gain factor is instantiated by the learned gains $G_{i,h,k}$ defined above.

\subsection{Full Layer Computation}\label{sec:FLC}

Consider a layer $\ell$ with $n_\ell$ input nodes and $n_{\ell+1}$ output nodes, and $K_{i,h}$ filters per edge $(i,h)$.

\paragraph{Edge function output.}
For an activation $a_i^{(\ell)}$ at node $i$ in layer $\ell$, the edge function from node $i$ to node $h$ in layer $\ell+1$ is
\begin{equation}
\Phi_{i,h}\bigl(a_i^{(\ell)}\bigr) = \sum_{k=1}^{K_{i,h}} H\!\left(f_{\mathrm{enc}}\bigl(a_i^{(\ell)}\bigr); \theta_{i,h,k}\right).
\end{equation}

\paragraph{Node aggregation.}
The pre-activation at node $h$ in layer $\ell+1$ is
\begin{equation}
y_h^{(\ell+1)} = \sum_{i=1}^{n_\ell} M_{i,h}^{(\ell)} \, \Phi_{i,h}\!\left(a_i^{(\ell)}\right),
\end{equation}
where $M_{i,h}^{(\ell)} \in \{0,1\}$ is an edge mask for layer $\ell$, used, for example, for pruning.
For the illustrative network in Fig.~\ref{fig:schematic_overview}b, $M_{i,h}^{(\ell)}$ is obtained by hard pruning during training, typically at each validation evaluation: for each edge $(i,h)$ in layer $\ell$, we evaluate the learned edge response at 1000 uniformly spaced points on the interval $[0,1]$, and set $M_{i,h}^{(\ell)}=0$ if the mean absolute response over those samples falls below a fixed pruning threshold of 0.05; otherwise $M_{i,h}^{(\ell)}=1$. Pruning is applied to the hidden layers only; edges in the output layer are always retained.

\paragraph{Inter-layer activation.}
For hidden layers, we apply the scaled sigmoid nonlinearity, constraining activations to values that encode to physically realisable frequencies:
\begin{equation}
a_h^{(\ell+1)} = \sigma_s\left(y_h^{(\ell+1)}\right)
\qquad \text{for hidden layers.}
\end{equation}

\subsubsection{Complete Network Forward Pass}

Given an input vector $\mathbf{x} = (x_1, \ldots, x_{n_0})$:

\begin{enumerate}
\item \textbf{Input transformation.} All inputs are bounded to the range [0,1]. For tasks where data may exceed these bounds, inputs are constrained with a sigmoid before passing to the PhyKAN:
\begin{equation}
a_i^{(0)} = \sigma_s(x_i),
\end{equation}
or, if thresholds $\tau_i$ are used,
\begin{equation}
a_i^{(0)} = \sigma_s(x_i - \tau_i).
\end{equation}
\item \textbf{Layer propagation.} For $\ell = 0, 1, \ldots, L - 1$,
\begin{equation}
y_h^{(\ell+1)} = \sum_{i=1}^{n_\ell} M_{i,h}^{(\ell)} \sum_{k=1}^{K_{i,h}}
G_{i,h,k}^{(\ell)}
H_{\text{BP}}\!\left(
        f_{\mathrm{enc}}(a_i^{(\ell)});
\nu^{\text{low}}_{i,h,k},
\nu^{\text{high}}_{i,h,k}
\right),
\end{equation}
where $H_{\text{BP}}$ denotes the band-pass magnitude response.

The activations in the next layer are
\begin{equation}
a_h^{(\ell+1)} =
\begin{cases}
\sigma_s\left(y_h^{(\ell+1)}\right), & \text{if } \ell < L - 1,\\[4pt]
y_h^{(\ell+1)}, & \text{if } \ell = L - 1.
\end{cases}
\end{equation}

\item \textbf{Output.} The network output is
\begin{equation}
\hat{y} = \mathbf{a}^{(L)}.
\end{equation}
\end{enumerate}
\subsection{Band-pass filters on programmable analogue hardware}\label{sec:experimental_measurements}
Input signals are sinusoidal voltage oscillations of $1\,\mathrm{V}_{\mathrm{RMS}}$ ($\sqrt{2}\,\mathrm{V}$ peak amplitude), with the input data magnitude encoded in the oscillation frequency, generated by an AIM-TTI TG1010A DDS function generator. Analogue filters were realised on an Anadigm AN231K04 development board operating on a 4 MHz clock frequency. This enables up to six band-pass filters, each realised as a cascade of first-order high-pass and low-pass stages, with corner frequencies between 4 and 400 kHz. The steady-state amplitude of each filter is measured by using a full-wave rectifier followed by a third-order low-pass filter with a corner frequency of 4 kHz. These voltages are summed using operational amplifiers with controllable gains, with the summed voltage read out using an NI BNC-2120 DAQ card.

As the hardware filters used here are based upon a switch-capacitor architecture, they behave as discrete-time filters. To account for the effects of frequency warping, a bilinear transform is used to map the frequencies of continuous filters ($\omega_a$) in equations described in Section~\ref{sec:transfunc} to frequencies at which discrete-time hardware equivalents have identical phase and gain ($\omega_d$) via Eq.~\ref{eq:freqwarp}:
\begin{equation}
\omega_d = \frac{2}{T} \cdot \arctan(\frac{{\omega_a T}}{2})
\label{eq:freqwarp}
\end{equation}
with $T = 2.5 \times 10^{-7} \,\text{s}$ representing the underlying clock period of the FPAA. This transformation is always applied to the input frequencies before the network forward pass to ensure parity between simulation and hardware.
\subsection{Hardware PhyKAN.} \label{sec:hardwarenetworks}
To generate a network in hardware, we first use the model to optimise parameter values to solve the task, and transfer the gain, low-pass corner frequency, and high-pass corner frequency to programmable analogue filters. As the development board used allows a maximum of six programmable filters per instance, a maximum of six sub-edges are used in each network edge for experimental demonstrations. To reconstruct an entire network from a single board, the response of the network is measured edge by edge, sampling the input--output relationship across the input range ($\approx$4.5--141 kHz) at logarithmically spaced steps over a 15\,s chirp sampled at a rate of 50 kS/s. The output is downsampled by a factor of 100 by decimating the signal twice, giving 7500 samples per edge. This gives a fine-grained, experimentally measured response function for each edge in the network. To simulate the response of a full network using experimentally measured data, linear interpolation is used to determine the predicted experimental response for task inputs from the fine-grained edge data gathered in the previous step.
\subsection{Hardware-Aware Training: Straight-Through Estimator}\label{sec:s_t_e}

Physical component values (e.g.\ filter corner frequencies and gains) take values only in a discrete set $\mathcal{D}\subset\mathbb{R}$. The gain is realised in discrete increments of 0.01, while available corner frequencies are stored in look-up tables available as part of the codebase (See Data Availability). To train such parameters, we use continuous updates while enforcing discrete values in the forward computation. In other words, optimisation uses a continuous surrogate for gradients, but the model behaves discretely when evaluated, following the straight-through estimator approach introduced by Bengio et al.~\cite{bengio2013estimating}.

Let $\theta\in\mathbb{R}$ be a trainable parameter and let
\[
T:\mathbb{R}\to\mathcal{D},\qquad \tilde{\theta}=T(\theta)
\]
denote the quantisation map selecting the physically realisable value. The map $T$ is piecewise constant, so
\[
\frac{dT(\theta)}{d\theta} = 0 \quad \text{except at jump discontinuities,}
\]
and at those jump points the derivative is undefined. Directly differentiating $\tilde{\theta}=T(\theta)$ would therefore yield zero or undefined gradients and stall optimisation.

The straight-through estimator is defined by separate forward and backward rules:
\begin{equation}
\tilde{\theta} \coloneqq T(\theta) \quad \text{in the forward pass},
\qquad
\left.\frac{\partial \tilde{\theta}}{\partial \theta}\right|_{\mathrm{STE}} \coloneqq 1
\quad \text{in the backward pass}.
\end{equation}
The forward computation therefore uses the physically realisable value, while the backward computation assigns the identity surrogate derivative in place of the derivative of $T$. For any loss $\mathcal{L}$ depending on $\tilde{\theta}$, this gives
\begin{equation}
\frac{\partial \mathcal{L}}{\partial \theta}
=
\frac{\partial \mathcal{L}}{\partial \tilde{\theta}}.
\end{equation}
\noindent
When a physical parameter is obtained from an unconstrained variable through a sigmoid reparameterisation, the sigmoid keeps the forward value within the admissible range and its derivative remains in the gradient through the chain rule.

After each continuous update of $\theta$, the next forward pass again uses the discretised value $T(\theta)$, so training explores a continuous parameter space while evaluation always reflects the discrete hardware constraints.
\subsection{Regularisation}\label{sec:regularisation}
To encourage sparsity, we add a regularisation term to the training objective.
\paragraph{L1 + entropy penalty.}
A sparsity-promoting term can be written as
\begin{equation}
\mathcal{L}_{\text{reg}} = \lambda \sum_{\ell}
\left[\frac{1}{N_\mu}\sum_{\mu=1}^{N_\mu}\sum_{i=1}^{N_i}\sum_{h=1}^{N_h}
|\Phi_{i,h}(a_{i,\mu}^{(\ell)})|
-
\sum_{i=1}^{N_i}\sum_{h=1}^{N_h} q_{i,h}^{(\ell)}
\log q_{i,h}^{(\ell)}
\right],
\end{equation}
where $\lambda$ is a regularisation coefficient, $\mu$ represents a sample index in a batch of size $N_\mu$, and $a_{i,\mu}^{(\ell)}$ is the activation of node $i$ in layer $\ell$ for sample $\mu$. We also define:
\begin{equation}
q_{i,h}^{(\ell)} \coloneqq \frac{1}{N_\mu}\sum_{\mu=1}^{N_\mu}\frac{|\Phi_{i,h}(a_{i,\mu}^{(\ell)})|}{\sum_i\sum_h|\Phi_{i,h}(a_{i,\mu}^{(\ell)})|}.
\end{equation}
\paragraph{Pruning/sparsification of networks.}
The regularisation term tends to suppress edge activations which carry correlated information. To remove redundant connections and reduce network size, the hard pruning mechanism outlined in Section~\ref{sec:FLC} was used. To prevent pruning networks too early while parameters adjust from initialisation, and to allow the network to fine-tune following removal of edges, pruning took place only between 10\% and 90\% of the total training iterations.
\subsection{Data generation for Six-axis Robot Kinematics}\label{sec:kinematics_data}
Robot kinematics are calculated for an IRB120 commercial robotic manipulator manufactured by ABB, which has been previously used for neural-network-based approximation of kinematics\cite{vsegota2021utilization}. Link lengths and rotational axes are defined according to the user manual, and are used to define Denavit--Hartenberg parameters for determining end-effector positions for known joint angles. This formulation recursively translates relative coordinate systems between joints $k$ and $k-1$ via transformation matrices $T_{k-1}^k$:

\indent$T_{k-1}^k =
\begin{bmatrix}
\mathrm{cos}\theta_k & -\mathrm{sin}\theta_k\mathrm{cos}\alpha_k & \mathrm{sin}\theta_k\mathrm{sin}\alpha_k & r_k\mathrm{cos}\theta_k \\
\mathrm{sin}\theta_k & \mathrm{cos}\theta_k\mathrm{cos}\alpha_k & -\mathrm{cos}\theta_k\mathrm{sin}\alpha_k & r_k\mathrm{sin}\theta_k \\
0 & \mathrm{sin}\alpha_k & \mathrm{cos}\alpha_k & d_k\\
0 & 0 & 0 & 1
\end{bmatrix}$

where:
\begin{itemize}
    \item $z^k$ lies along the rotational axis of joint $k+1$
    \item $x^k$ is normal to both $z^{k-1}$ and $z^k$ ($x^k=z^{k-1} \cross z^{k}$)
    \item $y^k$ is normal to the $x^kz^k$ plane such that it generates a right-handed coordinate system
    \item $\theta_k$ is the angle of joint $k$, about axis $z^{k-1}$ from $x^{k-1}$ to $x^k$
    \item $\alpha_k$ is the angle about axis $x^{k-1}$ from $z^{k-1}$ to $z^k$
    \item $r_k$ is the link length, the translational offset along axis $x^{k}$ from $z^{k-1}$ to $z^k$
    \item $d_k$ is the link offset, the translational offset along axis $z^{k-1}$ to the $x^kz^k$ plane
\end{itemize}
The robot consists of six revolute joints, meaning $\theta$ represents the only variable, here denoted by $\phi$, with $\alpha$, $r$, and $d$ remaining fixed for each $k$, defined by the robot geometry. The corresponding Denavit--Hartenberg parameters for each axis are defined in Table~\ref{tab:DH-params}. The end-effector location relative to the origin is found by calculating $T_0^6 = T_0^1 T_1^2 T_2^3 T_3^4 T_4^5 T_5^6$ for known joint angles [$\phi_1, ..., \phi_6$]. A dataset of 20,000 data points was generated by sampling each of the six joint angles independently and uniformly over the ranges below, and calculating the forward kinematics from the above equations. The sampled ranges are $\phi_1 \in [-165^\circ, 165^\circ]$, $\phi_2 \in [-110^\circ, 110^\circ]$, $\phi_3 \in [70^\circ, 110^\circ]$, $\phi_4 \in [-160^\circ, 160^\circ]$, $\phi_5 \in [-120^\circ, 120^\circ]$ and $\phi_6 \in [0^\circ, 360^\circ]$. Both MLP and PhyKAN networks receive inputs on the range [0,1], representing normalised joint angles relative to their minimum and maximum achievable positions.

Samples were split into training/validation/test sets consisting of 12000/3000/5000 samples respectively. The test samples were fixed as the final 5,000 samples in the generated dataset, before any shuffling. The specific training-validation splits were generated by randomly shuffling the remaining 15,000 samples according to a seed tied to the run number for each model, preserving specific shuffles across architectures for the independently trained runs.

For noisy training of the forward kinematics, the networks receive the ideal joint angles as inputs, while the training targets are end-effector locations obtained by evaluating the forward kinematics at joint angles perturbed by independent uniform noise of $\pm0.1$\,rad on each axis, emulating imperfect actuation. Test error is computed against the noiseless forward kinematics evaluated at the ideal joint angles.

For the inverse-kinematics feedforward controller, the kinematic equations together with the actuation noise constitute the ground-truth system, which the network observes only through the sampled measurements described above. The forward map is learned from those measurements, and the resulting forward model is frozen and used to close the training loop. The inverse network takes a target end-effector location as input and predicts joint angles; these are passed through the frozen forward model to obtain an estimated end-effector location, and the training loss is the mean-squared error between the target location and that estimate. Each inverse network is trained through a frozen forward model of its own architecture---PhyKAN inverse networks through a frozen PhyKAN forward model, MLP inverse networks through a frozen MLP forward model---both learned from the same measurements using identical seeds for data splitting. Test error is computed by evaluating the noiseless kinematic equations at the predicted joint angles and comparing the resulting location with the target.

\begin{table}[!h]
    \centering
    \begin{tabular}{|c|c|c|c|c|c|c|}
    \hline
         & $k=1$ & $k=2$ & $k=3$ & $k=4$ & $k=5$ & $k=6$\\
        \hline
        $\theta$ & $\phi_1$ & $\phi_2$ & $\phi_3$ & $\phi_4$ & $\phi_5$ & $\phi_6$\\
        $\alpha$ & $-\frac{\pi}{2}$ & 0 & -$\frac{\pi}{2}$ & $\frac{\pi}{2}$ & -$\frac{\pi}{2}$ & 0\\
        $r$ & 0 & 0.270 & 0.070 & 0 & 0.072 & 0\\
        $d$ & 0.290 & 0 & 0.134 & 0.168 & 0 & 0\\
    \hline
    \end{tabular}
    \caption{Denavit--Hartenberg parameters defining the six-axis robot. All lengths ($r$ and $d$) are given in metres, and all angles ($\theta$ and $\alpha$) are given in radians.}
    \label{tab:DH-params}
\end{table}

\subsection{Actor--Critic Training for Reinforcement Learning}\label{sec:AC_training}

The continuous-action experiments use a deterministic actor--critic update. The actor $\pi(s;\theta_A)$ maps a state $s$ to a continuous action, and the critic $Q(s,a;\theta_C)$ estimates the return, meaning the discounted future reward, associated with taking action $a$ in state $s$. During training, observed transitions $(s,a,r,s')$, consisting of the current state, action, reward, and next state, are stored in a replay buffer $\mathcal{D}$ and later sampled in mini-batches for parameter updates. This update is related to DDPG-style continuous-control methods~\cite{silver2014deterministic,lillicrap2015continuous}, but uses a reduced target-network structure: full DDPG uses target copies of both the actor and the critic, whereas here only the critic has a target copy. The output heads for all actor networks are linear and unbounded for both PhyKANs and MLPs.
\paragraph{Critic loss (TD learning).}
For a transition $(s,a,r,s') \sim \mathcal{D}$, the scalar value that the critic is trained to match is
\begin{equation}
y(s,a,r,s';\theta_A,\theta_{C'}) = r + \gamma \, Q'\bigl(s', \pi(s'; \theta_A); \theta_{C'}\bigr),
\end{equation}
where $\gamma$ is the discount factor and $Q'$ is the slowly updated target critic with parameters $\theta_{C'}$. The next action in this target is generated by the current actor, $\pi(s';\theta_A)$, and this action is then evaluated by the target critic.
The critic loss is then
\begin{equation}
\mathcal{L}_{\text{critic}}(\theta_C) =
\mathbb{E}_{(s,a,r,s') \sim \mathcal{D}}
\left[
\bigl(
Q(s, a; \theta_C) - y(s,a,r,s';\theta_A,\theta_{C'})
\bigr)^2
\right].
\end{equation}
During critic updates, $\theta_A$ and $\theta_{C'}$ are held fixed while $\theta_C$ is optimised.

\paragraph{Actor loss (deterministic policy gradient).}
The actor parameters $\theta_A$ are optimised to select actions that maximise the critic's estimate of long-term return. We define the actor objective
\begin{equation}
J_{\text{actor}}(\theta_A) =
\mathbb{E}_{s \sim \mathcal{D}_s}
\left[
Q\bigl(s, \pi(s; \theta_A); \theta_C\bigr)
\right],
\end{equation}
where $\mathcal{D}_s$ is the marginal state distribution induced by the replay buffer $\mathcal{D}$, and the critic parameters $\theta_C$ are treated as fixed.
In a loss-based formulation, the actor is trained to minimise
\begin{equation}
\mathcal{L}_{\text{actor}}(\theta_A) = - J_{\text{actor}}(\theta_A)
= - \mathbb{E}_{s \sim \mathcal{D}_s}
\left[
Q\bigl(s, \pi(s; \theta_A); \theta_C\bigr)
\right].
\end{equation}

\paragraph{Soft target update.}
To stabilise training, the target critic parameters are updated using a soft update rule
\begin{equation}
\theta_{C'} \leftarrow (1 - \tau_{\text{update}})\, \theta_{C'} + \tau_{\text{update}}\, \theta_C,
\end{equation}
where $\tau_{\text{update}} \in (0,1)$, here $\tau_{\text{update}}=0.001$ throughout, controls the update rate.
\paragraph{Task Hyperparameters}
For both the CartPole and Photovoltaic control tasks, both actor and critic parameters ($\theta_A$ and $\theta_C$ respectively) are optimised using an Adam optimiser. For both tasks, the filter-based PhyKANs were optimised with learning rates of $10^{-2}$ for the critic networks, and $10^{-3}$ for the actor networks. The memristor-based PhyKANs used the same learning rates except for the actor network for the Photovoltaic control, which used a learning rate of $10^{-4}$. For the MLP-based networks, learning rates of $10^{-3}$ were used for the critic networks and $10^{-5}$ for the actor networks in both tasks.

Complete training settings for every main-text figure are collected in Supplementary Note~\ref{sec:hyperparams}.

For all network architectures and tasks, replay buffers had a maximum size of 100,000 entries, which reset to the most recent 10,000 entries when the limit was exceeded during optimisation for the CartPole task. Training occurred after every step of the environment, randomly sampling 50 experiences from the replay buffer as a batch size for the optimisation process in all cases for actor and critic networks. Target networks (parameterised by $\theta_{C'}$) received a soft update after each step of the actor/critic networks according to the soft update rule outlined above. The CartPole and PV tasks used discount factors of 0.95 and 0.75 respectively for all architectures. Action selection was unbounded in all cases, and each actor deterministically mapped the current state to a continuous action for a given set of network parameters.

\subsection{CartPole Environment}\label{sec:RLmethods}
The CartPole control task environment follows the classic pole-balancing problem\cite{barto1983neuronlike} and was adapted from the OpenAI Gym library\cite{brockman2016openaigym}, with dynamic equations modified to accept a continuous force. The agent receives the state of the cart at every timestep $t$, $S_t$, defined by Eq.~\ref{eq:states}:
\begin{equation}
    S_t= [x_t, \dot{x}_t, \theta_t, \dot{\theta}_t]
    \label{eq:states}
\end{equation}
where $x_t$ denotes the position of the cart with respect to the centre of the environment at time $t$ in arbitrary units, $\dot{x}_t$ the velocity of the cart at time $t$ in arbitrary units per second, $\theta_t$ the angle of the pole with respect to the normal of the cart at time $t$ in radians, and $\dot{\theta}_t$ the angular velocity of the pole at time $t$ in radians per second. The agent provides a vector force along the x axis as output (as defined in Fig.~\ref{fig:CartPole}a), and is presented with a reward of +1 for every timestep for which it keeps the pole upright.
Rewards were shaped according to state observations to improve learning. The reward shaping function is defined by Eq.~\ref{eq:reward_shaping}:
\begin{equation}
    R^s_t=R_t-|x_t|-0.1\cdot |\dot{x}_t| - |\theta_t| - 0.5\cdot |\dot{\theta}_t|
    \label{eq:reward_shaping}
\end{equation}
where $R^s_t$ defines the shaped reward at time $t$, and $R_t$ the reward given by the environment at time $t$.
Networks were optimised over 1000 episodes with different random initial conditions, and trained according to the actor--critic paradigm outlined above.
\subsection{Data generation of Simulated Photovoltaic Array}\label{sec:PV_data}
Simulations of a photovoltaic array consisting of four individual modules were performed using a modified version of a Simulink model\cite{PV_model} of four connected SunPower SPR-X20-250-BLK modules operating at a fixed temperature of $25^\circ$C. Reference data were generated from 2000 random samples of irradiance, uniformly sampled in the range of $400$--$1000$ $\mathrm{W/m^2}$. Control voltage is ramped between $0$--$200$ V in steps of $0.1$ V, and simulated current and power generated by the array are recorded, generating simulation predictions for the power versus voltage response of the array under partial shading conditions.

These simulations form the basis for the environment in the PV array control task, acting as ground truth data for determining the outcome of agent actions. At the start of an episode, a random irradiance condition is sampled, as well as an initial control voltage. From the associated simulation data for the sampled irradiance conditions, the initial control voltage is used to determine the current, and consequently the power, generated by the array. The state vector provided to the actor and critic networks at time $t=0$ is given by the initial control voltage, $V_0$, current $I_0$, and power $P_0$, as well as the change in voltage $\Delta V_0=0$ on initialisation.
From this input state vector, the actor network produces a change in the control voltage, $\Delta V_t$, as its output, and the control voltage at the next timestep is given by $V_{t+1}=V_t+\Delta V_t$. The simulation data are consulted to give $I_{t+1}$ and $P_{t+1}$. These new state data are fed into the actor network, and the process occurs recursively until the fiftieth timestep, when the episode ends and a new irradiance condition is sampled. The filter-based PhyKAN, MemKAN, and MLP agents used here were trained over 2000 episodes, using a reward shaping function similar to that described in \cite{phan2020deep}:
\begin{equation}
    r = r_1 + r_2 + r_3 \\
\end{equation}
\begin{equation}
    r_1 = \frac{P_{t+1}}{1000} \\
\end{equation}
\begin{equation}
    r_2 = \begin{cases}
        r_1^2 & \text{if } P_{t+1}> P_t,\\
        0 & \text{otherwise}.\\
    \end{cases}
\end{equation}
\begin{equation}
    r_3 = \begin{cases}
        0 & \text{if } 0 < V_{t+1} < 200,\\
        -1 & \text{otherwise.}
    \end{cases}
\end{equation}
For the MLP controls, the magnitude of the out-of-bounds penalty $r_3$ was increased tenfold, from $-1$ to $-10$, because the $-1$ penalty did not sufficiently discourage voltage excursions beyond the permitted range during training, resulting in significantly worse performance than quoted in the figures which used the stronger penalty. In all cases where controller-imposed voltage exceeds the $0-200$ V range, voltages are clipped to the appropriate bound.
\subsection{Analytical Modelling of Memristor Circuits}\label{sec:memdiode_model}

The memristors studied here are two-terminal devices whose current--voltage (I/V) relationship depends on an internal state variable, enabling a history-dependent response to applied stimuli \cite{Chua1971,Strukov2008}. A defining feature of such devices is their ability to switch between distinct resistance states, typically referred to as the low-resistance state (LRS) and the high-resistance state (HRS), as shown in Supplementary Fig.~\ref{fig:memristor_IV}, each associated with a different I/V characteristic. This behaviour arises from the coupling between electronic transport and the spatial distribution of defects within the active material, typically governed by nanoscale redox processes \cite{Waser2009,Valov2013}.

In filamentary devices, this resistive switching is commonly associated with the formation and rupture of a conductive filament bridging the electrodes (Supplementary Fig.~\ref{fig:memristor_physics}). The LRS corresponds to a continuous filament that provides a quasi-metallic conduction path, whereas the HRS is characterised by a partially ruptured filament, where a nanoscale gap separates the filament tip from the electrode. In this regime, conduction is dominated by barrier-limited transport mechanisms across the gap, giving rise to a strongly nonlinear, often exponential, current--voltage relationship \cite{Sze2006}. By applying appropriate voltage signals, the device can be reversibly tuned between these configurations, effectively controlling both the filament continuity and the associated transport mechanism. As a result, both the overall conductance and the degree of nonlinearity can be precisely adjusted through electrical programming, providing a versatile platform for analogue signal processing.

To model the memristor's behaviour in the proposed active circuit, the Dynamic Memdiode Model (DMM) for memristors is used, in which the device is represented by two coupled equations: an electronic transport equation and a memory-state equation \cite{Aguirre2022}. In this framework, the memristive device is modelled as a nonlinear conduction element whose characteristics depend on an internal state variable $\lambda$, which evolves dynamically under the applied electrical stimulus.

The electronic transport equation is expressed in terms of the voltage across the conductive constriction as

\begin{equation}
V_C = V - I R_I ,
\label{eq:constriction_voltage}
\end{equation}

where $V$ is the applied voltage, $I$ is the device current, and $R_I$ accounts for the series resistance associated with the filament structure. The current is then given by

\begin{equation}
I(V_C) = I_0(\lambda)\sinh\left[\alpha\left(V_C - R_S(\lambda)I\right)\right],
\label{eq:dmm_current}
\end{equation}
where $\alpha(\lambda)$ is a fitting parameter that controls the voltage sensitivity of the conduction process (the higher $\alpha$, the more nonlinear the conduction is), $R_S(\lambda)$ is a state-dependent resistance term, and $I_0(\lambda)$ is a state-dependent current scaling factor. This expression captures the nonlinear current--voltage behaviour and its dependence on the internal state of the device.

The internal state variable $\lambda$ is normalised between 0 and 1, representing the high-resistance state (HRS) and low-resistance state (LRS), respectively. The dependence of the current amplitude on the state variable is described as
\begin{equation}
I_0(\lambda) = \left(I_{\mathrm{on}} - I_{\mathrm{off}}\right)\lambda + I_{\mathrm{off}},
\label{eq:dmm_i0}
\end{equation}
where $I_{\mathrm{on}}$ and $I_{\mathrm{off}}$ define the current levels corresponding to the LRS and HRS limits. Similar dependencies exist for $\alpha(\lambda)$ and $R_S(\lambda)$, involving limiting parameters $\alpha_{min}$, $\alpha_{max}$, $R_{S_{min}}$, and $R_{S_{max}}$.

The evolution of the internal state is governed by
\begin{equation}
\frac{d\lambda}{dt} =
\frac{1 - \lambda}{\tau_S(\lambda, V_C)} -
\frac{\lambda}{\tau_R(\lambda, V_C)},
\label{eq:dmm_memory}
\end{equation}
where $\tau_S$ and $\tau_R$ are the characteristic SET and RESET times, respectively. These characteristic times depend on the voltage and are defined as
\begin{equation}
\tau_S(\lambda, V_C) = \tau_0
\exp\left[-\eta_S\left(V_C - V_S\right)\right],
\label{eq:dmm_tau_set}
\end{equation}
\begin{equation}
\tau_R(\lambda, V_C) = \tau_0
\exp\left[\eta_R \lambda^\gamma \left(V_C - V_R\right)\right],
\label{eq:dmm_tau_reset}
\end{equation}
where $\tau_0$ is a characteristic time constant that sets the units, $\eta_S$ and $\eta_R$ are fitting parameters that control the voltage dependence of the transition rates, $V_S$ and $V_R$ are the SET and RESET voltages, and $\gamma$ accounts for nonlinear effects during RESET.

To reproduce snapback behaviour during SET, the model allows the SET voltage to depend on the current as
\begin{equation}
V_S(I) =
\begin{cases}
V_T, & I \geq I_{\mathrm{SB}}, \\
V_{S0}, & I < I_{\mathrm{SB}},
\end{cases}
\label{eq:dmm_snapback}
\end{equation}
where $V_T$ is the transition voltage, $V_{S0}$ is the low-current SET voltage, and $I_{\mathrm{SB}}$ is the snapback current threshold. The SET voltage $V_S$ appearing in Eq.~\ref{eq:dmm_tau_set} is this current-dependent quantity.

The coupling between the transport equation and the state equation enables the model to reproduce the nonlinear and history-dependent behaviour characteristic of bipolar resistive switching devices. This model can also be synthesised in an equivalent circuit model, as shown in Supplementary Fig.~\ref{fig:memristor_model}, making it suitable for SPICE-like simulation tools.

Based on this model, the proposed active synapse is depicted in Fig.~\ref{fig:MemKAN}a. The transfer function for each case is given by Eq.~\ref{eq:BP_transfer}.

\begin{equation}
V_{out}(x) = R_L \left( \ \frac{V_{in} - V_{out}}{R} + I_0 e^{\alpha\left[\left(V_{in} - V_{out} \right) - R_S I_{\mathrm{MR}} \right]}\right)
\label{eq:BP_transfer}
\end{equation}
where $I_{\mathrm{MR}}$ is the current through the memristor, that is, the second term in the bracket. For this transfer function, the fitting parameters used to tune the synapse response are $R_L$ (Supplementary Fig.~\ref{fig:RL_BP_impact}), $R$ (Supplementary Fig.~\ref{fig:R_BP_impact}), $R_S$ (Supplementary Fig.~\ref{fig:RS_BP_impact}), $\alpha$ (Supplementary Fig.~\ref{fig:alpha_BP_impact}) and $I_0$ (Supplementary Fig.~\ref{fig:Io_BP_impact}). The first two are circuit parameters; the remaining three are fitting parameters of the memristor model. Although treated as three independent fitting parameters, these could also be represented as functions of $\lambda$, according to Eq.~\ref{eq:dmm_i0} in the scenario of a memristor whose state is tuneable and can achieve different nonlinear HRS states as $\lambda$ goes from 0 to 1.
\subsection{Multilayer Perceptron Models of Memristor Circuits}\label{sec:memristor_MLP}
To reduce the computational cost of backpropagation through the nonlinear-least-squares solver used to find $V_{out}$ in Eq.~\ref{eq:BP_transfer} for the memristor circuit, a data-driven approach was taken. First, data were generated via the analytical model, consisting of 20,000 samples of the five control parameters, with 200 input voltages randomly sampled per set of control parameters to give 4 million data points.

An MLP of shape [6, 125, 125, 1] was trained to interpolate the memristor's transfer function with respect to voltage input and the control parameters over a training set of 3 million data points, with a validation set of 500,000 data points used to determine hyperparameters such as network size and learning rate. The model converged with a mean-squared error of $5.17\times10^{-7}$ evaluated on an unseen dataset consisting of the remaining 500,000 data points. Details of hyperparameter optimisation can be found in Supplementary Note~\ref{sec:hyperparamMLP}.

\subsection{Memristor-based PhyKANs}\label{MemPhyKAN}
The memristor-based networks are constructed similarly to the filter-based PhyKANs described above. Learnable parameters on the network edges are the five control parameters that dictate the memristor circuit response, $\theta_{i,h,k}$, a bias on inputs provided to the edge $b_{i,h,k}$, plus an analogue gain $g_{i,h,k}$ applied after the circuit. No frequency-encoding function was used; however, the sigmoid activation that bounds input ranges and activations at nodes in hidden layers remained. The response of edge functions is given by a forward pass through the MLP, which takes activations and learnable parameters $\theta$ as inputs. The parameters of the MLP, $W$, remain fixed and identical for all nodes. Edge functions, node aggregation, and node activations are as follows:
\begin{equation}
    \theta_{i,h,k}=(R_{L,i,h,k}, R_{i,h,k}, R_{S,i,h,k}, \alpha_{i,h,k}, I_{0,i,h,k}),
\end{equation}
\begin{equation}
\Phi_{i,h}\bigl(a_i^{(\ell)}\bigr) = \sum_{k=1}^{K_{i,h}} g_{i,h,k}\cdot H_{\mathrm{MLP}}\!\left((a_i^{(\ell)}-b_{i,h,k}, \theta_{i,h,k});W\right),
\end{equation}
\begin{equation}
y_h^{(\ell+1)} = \sum_{i=1}^{n_\ell} M_{i,h}^{(\ell)} \sum_{k=1}^{K_{i,h}}
g_{i,h,k}\cdot H_{\mathrm{MLP}}\!\left((a_i^{(\ell)}-b_{i,h,k},
\theta_{i,h,k});W\right),
\end{equation}
\begin{equation}
a_h^{(\ell+1)} =
\begin{cases}
\sigma_s\left(y_h^{(\ell+1)}\right), & \text{if } \ell < L - 1,\\[4pt]
y_h^{(\ell+1)}, & \text{if } \ell = L - 1.
\end{cases}
\end{equation}

\subsection{Power Consumption Estimations}
\label{sec:power_cons}
The circuitry used in the experiments was chosen for reconfigurability and ease of measurement. For low-power edge-computing applications, the switch-capacitor-based FPAA used for network edges and the DDS arbitrary function generator used on network nodes consume considerably more power than would be expected on a dedicated ASIC. The AN231E04 device used on the experimental AN231K04 development board is specified to draw typically 40 mA at 3.3 V (approximately 132 mW), although the manufacturer notes that current after configuration depends strongly on the programmed circuit \cite{anadigm2013an231k04}. We therefore estimate the potential power savings of a well-optimised PhyKAN platform using plausible functionally equivalent components from the literature. As these components are implemented in a variety of technologies and operate at different speeds, we extrapolate from published figures to obtain a realistic estimate.

As some calculations here are based on theoretical constraints, the numbers should be treated as lower-bound estimates for a manufactured platform, and should therefore be read as an indication of the achievable power range, not as a design specification. The proposed architecture is designed around a 1 V supply voltage. Network edges used in our demonstration require a parallel bank of six band-pass filters---each itself a cascaded high-pass/low-pass pair---with its own peak-detection circuit and amplifier. Each of the two stages within a band-pass filter contributes one 1 pF capacitor, so a band-pass filter presents 2 pF and the six-filter bank forming one edge presents 12 pF; these three figures recur below and all follow from the single 1 pF choice.

To realise the filters, we propose sub-threshold transconductance amplifiers, and 1 pF capacitors. By lowering operating frequency closer to audio frequency range (100 Hz--50 kHz), the operating power of the band-pass filter stage can be further reduced. To achieve 50 kHz corner frequencies with a 1 pF capacitor, transconductance values of $G_m = 2\pi f_c C = 314$ nS are required for $V_{DD}$ = 1 V. For an overdrive voltage of 200 mV, this gives a required bias current of 31.4 nA ($I_{\mathrm{bias}} = G_m \cdot V_{\mathrm{ov}}/2$). Employing a folded cascode configuration to reduce noise on the amplifiers, we can assume a doubling of the required bias currents, giving a total of 62.8 nW power consumption per single filter stage, or 753.6 nW for the filter bank of six band-pass filters (12 overall stages). In existing works, programmable band-pass filters based upon operational transconductance amplifiers have been realised in 350 nm CMOS with power consumptions of 1.31 $\mu$W and frequency tuning ratios greater than 10,000 \cite{peng2017power}. For fixed filters based upon 180 nm CMOS, powers as low as 41 nW have been realised \cite{lee2018low}.

To extract the amplitude of the filtered signal and provide analogue gains to the outputs, additional components are required. To measure the amplitude of the signal, we propose a source follower stage to track the envelope of the oscillations. To capture 50 kHz oscillations using a comparator-based rectifier, a 100 kHz gain-bandwidth product is required. For a PMOS comparator with a gate capacitance of 50 fF, the required transconductance is given by $G_m=2\pi \cdot GBW \cdot C_{\mathrm{gate}}$ = 31 nS. Assuming weak inversion (sub-threshold regime), this leads to a bias current of $I_{\mathrm{bias}} = G_m \cdot nV_T = 1.2$ nA or 1.2 nW of power at $V_{DD} = 1$ V. To amplify the envelope signal, a low gain-bandwidth amplifier operating at $\approx$100 Hz can be used, requiring bias currents of ${\sim}1$ nA and powers of 1 nW. In total, the peak detection and amplification stages require 2.2 nW per band-pass filter, or 13.2 nW for the filter bank. This brings the total power for each edge to 766.8 nW.

Network nodes are required to convert the amplified edge responses into frequency-encoded oscillations for the subsequent layer of edges. Using a relaxation oscillator and a differential pair shaper to approximate a sine wave, a range of 10 Hz to 50 kHz can be covered, spanning the operating band assumed above. Assuming the highest input frequency and 1 V peak-to-peak amplitude, the oscillator requires a slew rate of 157 kV/s ($0.5 \cdot 2\pi \cdot 50{,}000$). The node drives the 12 pF presented by each connected filter bank, so its load, and hence its power, scales with the number of edges it feeds; the oscillator cost is therefore quoted per driven edge. At 12 pF this gives $I = C\,\mathrm{SR} = 1.88\;\mathrm{\mu A}$, or 1.88 $\mathrm{\mu}$W for signal generation, bringing the total to 2.65 $\mathrm{\mu}$W per edge in the network. In existing works, oscillators are reported with per-cycle power efficiencies ranging between 1.9 nW/kHz in 350 nm CMOS\cite{denier2010analysis}, to 0.68 nW/kHz in 40 nm CMOS\cite{8796364}.

For the PV control task as shown in Fig.~\ref{fig:PV_Control}, the smallest networks (after pruning) to achieve power generation ratios above 0.9 required 13 edges in the network, leading to a total estimated power consumption of 34.4 $\mathrm{\mu}$W.  This is one to two orders of magnitude below microcontrollers capable of running networks with enough parameters to match PhyKAN performance. For example, on the MLPerf Tiny benchmark \cite{banbury2021mlperf}, microcontrollers capable of running 48k parameter neural networks consume between 0.33 mW (Syntiant NDP9120, 21 MHz clock) and 21.56 mW (STM NUCLEO-L4R5ZI-P, 70 MHz clock) during inference. To benchmark the throughput of the device, we calculate the time taken for the slowest filters to settle. For 4 kHz corner frequencies, the filters have a time-constant $\tau$ of $\tau = \frac{1}{2\pi f_c} \approx 40\mu s$. To settle within $0.1\%$ of the steady-state level takes 7 time-constants, or 0.28 ms, enabling a throughput of around 3.5 kHz. The component-level assumptions behind this estimate are summarised in Table~\ref{tab:power-cons}, where each value quoted is for a single filter (sub-edge) in the network.

\begin{table}[!hpt]
    \centering
    \begin{tabular}{|l|c|c|c|c|}
    \hline
         & \shortstack{Theoretical\\Design} & \shortstack{Literature\\Examples} & \shortstack{CMOS\\process} & Comments\\
        \hline
        \shortstack[l]{Band-pass\\Filter} & $125.6 \mathrm{nW}$ & \shortstack{$1.31 \mathrm{\mu W}$\cite{peng2017power}\\$41 \mathrm{nW}$\cite{lee2018low}} & \shortstack{$350 \mathrm{nm}$\\$180 \mathrm{nm}$} & \shortstack{$20 \mathrm{kHz}$\\$250 \mathrm{Hz}$}\\
        \hline
        \shortstack[l]{Signal\\Generation} & $6.28 \mathrm{nW/kHz}$ & \shortstack{$1.9 \mathrm{nW/kHz}$\cite{denier2010analysis}\\$0.68 \mathrm{nW/kHz}$\cite{8796364}} & \shortstack{$350 \mathrm{nm}$\\$40 \mathrm{nm}$} & \shortstack{2 pF load per band-pass\\filter (2 stages, 1 pF each)}\\
        \hline
        \shortstack[l]{Envelope\\tracking} & $1.2 \mathrm{nW}$ & - & - & Max 50 kHz\\
        \hline
        \shortstack[l]{DC\\Amplification} & $1.0 \mathrm{nW}$ & - & - & \shortstack{DC level changes\\$\approx 100 \mathrm{Hz}$}\\
    \hline
    \end{tabular}
    \caption{Theoretical design power consumption and literature examples for each stage required to create low-power PhyKAN edges. CMOS processes refer to the literature citations.}
    \label{tab:power-cons}
\end{table}

\FloatBarrier
\section*{Data Availability}
All experimental data, trained models, and raw data used to generate plots can be downloaded from \url{https://doi.org/10.6084/m9.figshare.33955060}

\section*{Code Availability}
All scripts used to generate results for this paper can be found at \url{https://github.com/ianvidamour/PhyKAN}. The code used to recreate the figures from extracted data can be found at \url{https://github.com/eleni-vasilaki/PhyKAN}.

\section*{Competing Interests}
The authors declare the following competing interests: The University of Sheffield has filed European Patent Application No. 26387088.3, ``Nonlinear functions on network connections'', relating to the physical-KAN architecture and analogue-filter implementation described in this work. E.V. and I.T.V. are named inventors. The application is pending. The remaining authors declare no competing interests.

\clearpage
\appendix

\setlength{\headheight}{24pt}
\pagestyle{fancy}
\fancyhf{}
\fancyhead[C]{{\bfseries Supplementary Material:}~\shorttitle}
\fancyfoot[C]{\thepage}

\renewcommand{\headrulewidth}{0.4pt}

\renewcommand{\thesection}{S\arabic{section}}
\renewcommand{\thefigure}{S\arabic{figure}}
\renewcommand{\thetable}{S\arabic{table}}
\renewcommand{\theequation}{S\arabic{equation}}

\renewcommand{\theHsection}{supp.\arabic{section}}
\renewcommand{\theHfigure}{supp.\arabic{figure}}
\renewcommand{\theHtable}{supp.\arabic{table}}
\renewcommand{\theHequation}{supp.\arabic{equation}}

\setcounter{section}{0}
\setcounter{figure}{0}
\setcounter{table}{0}
\setcounter{equation}{0}

\setcounter{page}{1}
\renewcommand{\thepage}{S\arabic{page}}

\thispagestyle{fancy}

\section{Training Hyperparameters}\label{sec:hyperparams}

The tables below give the network and training settings used to produce each main-text figure.

\begin{table}[!h]
    \centering
    \begin{tabular}{|l|c|c|}
    \hline
        & \textbf{Panel b demonstration} & \textbf{Panel c sweep}\\
    \hline
        Architecture & $[2,3,2,1]$ & $[2,5,1]$ to $[2,5,5,5,5,1]$\\
        Filter units per edge & 6 & 6\\
        Optimiser & Adam & Adam\\
        Learning rate & $10^{-3}$, constant & $10^{-2}$, $\times 0.99$ per pruning step\\
        Training iterations & 60{,}000 & 200{,}000\\
        Batch size & 2{,}000 & 2{,}500\\
        Regularisation $\lambda$ & $10^{-4}$ throughout & $10^{-4}$, then $10^{-9}$ after 150{,}000\\
        Pruning threshold & 0.1 & 0.05\\
        Pruning schedule & every 5{,}000 over 20{,}000--50{,}000 & every 5{,}000 over 20{,}000--150{,}000\\
        Training data & uniform on $[0,1]^2$, on the fly & 80{,}000 train / 20{,}000 test\\
        Seed & 0 & --\\
    \hline
    \end{tabular}
    \caption{Training settings for the Feynman networks in Fig.~\ref{fig:schematic_overview}.
    The panel b network approximates $\cos(X_0) + X_1\cos^2(X_0)$ on $[0,1]^2$. The panel c
    sweep covers 27 dimensionless Feynman functions at four depths, $[2,5,1]$, $[2,5,5,1]$,
    $[2,5,5,5,1]$ and $[2,5,5,5,5,1]$. In the panel c
    sweep the learning-rate scheduler is stepped only when pruning occurs, so it fires 25 times
    over the run.}
    \label{tab:hyper-feynman}
\end{table}

\begin{table}[!h]
    \centering
    \resizebox{\linewidth}{!}{%
    \begin{tabular}{|l|c|c|c|c|}
    \hline
        & \multicolumn{2}{c|}{\textbf{Forward}} & \multicolumn{2}{c|}{\textbf{Inverse}}\\
        & PhyKAN & MLP & PhyKAN & MLP\\
    \hline
        Architecture (1 hidden layer) & $[6,N,3]$ & $[6,N,3]$ & $[3,N,6]$ & $[3,N,6]$\\
        Architecture (2 hidden layers) & $[6,N,N,3]$ & $[6,N,N,3]$ & $[3,N,N,6]$ & $[3,N,N,6]$\\
        Filter units per edge & 6 & -- & 6 & --\\
        Optimiser & Adam & Adam & Adam & Adam\\
        Learning rate & $10^{-2}$ & $10^{-3}$ & $10^{-2}$ & $10^{-4}$\\
        Scheduler & $\times 0.99$ / 1{,}000 & none & $\times 0.99$ / 1{,}000 & none\\
        Batch size & 500 & 500 & 500 & 500\\
        Updates & 100{,}000 & 100{,}000 & 50{,}000 & 50{,}000 (1L)/ 30{,}000 (2L)\\
        Validation interval & 1{,}000 & 2{,}000 & 1{,}000 & 1{,}000\\
        Regularisation $\lambda$ & $10^{-5}$ & -- & $10^{-5}$, then $10^{-9}$ & --\\
        Weight decay & -- & -- & -- & $10^{-5}$ (2 layers)\\
        Hardware-discrete phase & final 20{,}000 & -- & final 20{,}000 & --\\
        Pruning threshold & -- & -- & $10^{-3}$ & --\\
        Pruning schedule & -- & -- & every 2{,}000 over 20{,}000--30{,}000 & --\\
        Early stopping & none & none & none & counter $\geq 3$\\
    \hline
    \end{tabular}}
    \caption{Training settings for the six-axis kinematics networks in
    Fig.~\ref{fig:Robotics}. The dataset is 20{,}000 samples split 12{,}000 / 3{,}000 / 5{,}000
    into training, validation and test, with the training--validation shuffle seeded by the run
    index. Forward targets are end-effector locations evaluated at joint angles perturbed by
    independent uniform noise of $\pm 0.1$\,rad per axis. Each inverse network is trained through
    a frozen forward model of its own architecture.}
    \label{tab:hyper-kinematics}
\end{table}

\begin{table}[!h]
    \centering
    \begin{tabular}{|l|c|c|}
    \hline
        & \textbf{PhyKAN} & \textbf{MLP}\\
    \hline
        Actor (2 hidden layers) & $[4,N,N,1]$ & $[4,N,N,1]$\\
        Critic (2 hidden layers) & $[5,N,N,1]$ & $[5,N,N,1]$\\
        Actor (3 hidden layers) & $[4,N,N,N,1]$ & $[4,N,N,N,1]$\\
        Critic (3 hidden layers) & $[5,N,N,N,1]$ & $[5,N,N,N,1]$\\
        Filter units per edge & 6 & --\\
        Hidden activation & -- & ReLU\\
        Actor learning rate & $10^{-3}$ & $10^{-5}$\\
        Critic learning rate & $10^{-2}$ & $10^{-3}$\\
        Episodes & \multicolumn{2}{c|}{1{,}000}\\
        Replay buffer & \multicolumn{2}{c|}{100{,}000, reset to 10{,}000}\\
        Replay batch & \multicolumn{2}{c|}{50}\\
        Target update $\tau$ & \multicolumn{2}{c|}{0.001}\\
        Discount factor & \multicolumn{2}{c|}{0.95}\\
    \hline
    \end{tabular}
    \caption{Network and training settings for the continuous-action CartPole agents in
    Fig.~\ref{fig:CartPole}.}
    \label{tab:hyper-cartpole}
\end{table}

\begin{table}[!h]
    \centering
    \begin{tabular}{|l|c|c|}
    \hline
        & \textbf{PhyKAN} & \textbf{MLP}\\
    \hline
        Actor & $[4,N,1]$ & $[4,N,N,1]$\\
        Critic & $[5,N,1]$ & $[5,N,N,1]$\\
        Filter units per edge & 6 & --\\
        Hidden activation & -- & sigmoid\\
        Actor learning rate & $10^{-3}$ & $10^{-5}$\\
        Critic learning rate & $10^{-2}$ & $10^{-3}$\\
        Critic regularisation $\lambda$ & $10^{-4}$ & --\\
        Action penalty coefficient & \multicolumn{2}{c|}{$10^{-4}$ on $|a|$}\\
        Episodes $\times$ timesteps & \multicolumn{2}{c|}{$2{,}000 \times 50$}\\
        Discount factor & \multicolumn{2}{c|}{0.75}\\
        Target update $\tau$ & \multicolumn{2}{c|}{0.001}\\
        Exploration $\epsilon$ & \multicolumn{2}{c|}{0.33}\\
        Out-of-bounds penalty $r_3$ & $-1$ & $-10$\\
    \hline
    \end{tabular}
    \caption{Network and training settings for the photovoltaic control agents in
    Fig.~\ref{fig:PV_Control}. The state supplied to both networks is
    $[V_t,\ I_t,\ \Delta V_t,\ P_t/1000]$, with the voltage divided by 200 and the current by 9
    before entering the network; the voltage change and the power ratio are passed unscaled.}
    \label{tab:hyper-pv}
\end{table}

\begin{table}[!h]
    \centering
    \begin{tabular}{|l|c|}
    \hline
        Architecture & $[6,N,N,3]$\\
        Memristor units per edge & 3\\
        Optimiser & Adam\\
        Learning rate & $10^{-2}$, $\times 0.99$ per 1{,}000 updates\\
        Updates & 100{,}000\\
        Batch size & 500\\
        Regularisation $\lambda$ & $10^{-5}$\\
        Validation interval & 1{,}000\\
        Surrogate MLP & $[6,125,125,1]$\\
    \hline
    \end{tabular}
    \caption{Training settings for the memristor-based PhyKANs in Fig.~\ref{fig:MemKAN}. Each
    edge carries three memristor units, and each unit has five control parameters, an input bias
    and an output gain. The surrogate MLP that supplies the edge response is trained separately
    and held fixed.}
    \label{tab:hyper-memkan}
\end{table}

\FloatBarrier

\section{Universality Argument for Idealised PhyKANs}\label{sec:uat}

\medskip
\noindent\textbf{Statement.} For every continuous function \(F\) on a compact domain \(Q=I_1\times\cdots\times I_d\subset\mathbb{R}^d\) and every \(\varepsilon>0\), there exists an idealised PhyKAN network \(F_{\rm P}\) such that \(\sup_{x\in Q}|F(x)-F_{\rm P}(x)|<\varepsilon\).
\medskip

This note proves that idealised PhyKAN networks are universal approximators on compact domains. The proof has five parts.
First, we show that an equal-corner band-pass filter gives a translated bump, and that changing the corner frequencies moves the centre of the bump. A PhyKAN edge is a finite signed sum of such translated bumps, because each filter in the bank has its own gain.
Second, we show that finite sums of translated bumps can approximate any sine or cosine on a compact interval.
Third, Stone--Weierstrass is applied to the algebra generated by the constant function, sines, and cosines.
Fourth, a single idealised PhyKAN edge is a one-dimensional universal approximator.
Finally, we show that a PhyKAN network is a universal function approximator via the standard sigmoid universal approximation theorem: the weighted sums in a classical ANN can be replaced by one-dimensional PhyKAN edge approximators.

\medskip
\noindent
Throughout, expressions of the form
\[
\sup_{x\in K}|p(x)-q(x)|
\]
use \(\sup\) for the supremum, which is the largest approximation error
between \(p\) and \(q\) over the compact set \(K\). This quantity is the
worst-case error on \(K\). Thus a bound such as
\[
\sup_{x\in K}|p(x)-q(x)|<\delta
\]
means that \(|p(x)-q(x)|<\delta\) for every \(x\in K\).

\subsection{Equal-corner filters give translated bumps}

An input value \(x\) is not applied to the filter directly; it is encoded as
the driving frequency
\[
\begin{aligned}
f_{\rm enc}(x)
&=10^{\alpha_{\text{in}}+\beta_{\text{in}} x}\\
&=e^{(\alpha_{\text{in}}+\beta_{\text{in}} x)\ln 10}\\
&=e^{a+\kappa x},\\
a&:=\alpha_{\text{in}}\ln 10,\quad \kappa:=\beta_{\text{in}}\ln 10>0.
\end{aligned}
\]
Thus we are using the same log-frequency encoding as before, only written in
base \(e\).

A single band-pass filter is the cascade of a first-order high-pass stage and
a first-order low-pass stage. Let \(\nu_{\rm HP}\) denote the high-pass corner
frequency and \(\nu_{\rm LP}\) the low-pass corner frequency. Both are tuneable
filter parameters. To obtain a translated bump centred at the input location
\(c\), tune both corners to the frequency produced when the input is \(c\):
\[
\nu_{\rm HP}(c)=\nu_{\rm LP}(c)=:\nu_c=f_{\rm enc}(c).
\]
Here \(c\) is the chosen centre of this filter response in the input coordinate. Changing \(c\) retunes the two
corner frequencies and translates the bump along the input axis.

For a first-order high-pass stage and a first-order low-pass stage with the
same corner \(\nu_c\), the magnitude responses are
\[
|H_{\rm HP}(\nu;\nu_c)|
=
\frac{\nu/\nu_c}{\sqrt{1+(\nu/\nu_c)^2}},
\qquad
|H_{\rm LP}(\nu;\nu_c)|
=
\frac{1}{\sqrt{1+(\nu/\nu_c)^2}}.
\]
These are the standard first-order filter magnitude responses, and cascaded
linear time-invariant transfer functions multiply\cite{oppenheim1997signals}.
The product rule assumes that the two stages are buffered, or equivalently
that inter-stage loading is negligible.
Their cascade has magnitude equal to the product, hence
\[
|H_{\rm HP}(\nu;\nu_c)|\,|H_{\rm LP}(\nu;\nu_c)|
=
\frac{\nu/\nu_c}{1+(\nu/\nu_c)^2}.
\]
Driving the filter at \(\nu=f_{\rm enc}(x)\) and setting
\(\nu_c=f_{\rm enc}(c)\), the ratio is
\[
\frac{\nu}{\nu_c}
=
\frac{e^{a+\kappa x}}{e^{a+\kappa c}}
=
e^{\kappa(x-c)}.
\]
The unnormalised equal-corner response is therefore the cascade magnitude
evaluated at these encoded frequencies:
\[
\begin{aligned}
|H_{\rm HP}(f_{\rm enc}(x);f_{\rm enc}(c))|\,
|H_{\rm LP}(f_{\rm enc}(x);f_{\rm enc}(c))| =
\frac{e^{\kappa(x-c)}}{1+e^{2\kappa(x-c)}}.
\end{aligned}
\]
Its value at the centre \(x=c\) is \(1/2\), so after normalising the peak to
one we obtain
\[
B(x-c)
:=
\frac{2e^{\kappa(x-c)}}{1+e^{2\kappa(x-c)}}
=
\frac{2}{e^{\kappa(x-c)}+e^{-\kappa(x-c)}}
=
\operatorname{sech}\big(\kappa(x-c)\big).
\]
Here \(\operatorname{sech}\) denotes the hyperbolic secant,
\[
\operatorname{sech} z=\frac{1}{\cosh z}=\frac{2}{e^z+e^{-z}}.
\]

A PhyKAN edge is a bank of such filters whose outputs are added with
trainable gains. Consequently, define the idealised edge family on the real
line by
\[
\mathcal F_{\rm edge}(\mathbb{R})
:=
\left\{
x\mapsto \sum_{k=1}^{K}G_k B(x-c_k)
:\ K<\infty,\ G_k,c_k\in\mathbb{R}
\right\}.
\]
These sums are defined for every real input \(x\). When the approximation
problem is posed on a compact interval \(I\subset\mathbb{R}\), we use the same
sums but evaluate them only for \(x\in I\). We denote this interval-restricted
use by \(\mathcal F_{\rm edge}(I)\).

\subsection{Scaled and translated bumps approximate trigonometric polynomials}

From the previous subsection, an idealised PhyKAN edge has the form
\[
\sum_{k=1}^{K}G_k B(x-c_k).
\]
We first consider the
continuous analogue
\[
S_G(x)=\int_{\mathbb R}G(c)B(x-c)\,dc,
\]
where \(G(c)\) is the gain assigned to the bump centred at \(c\). To build a
cosine with frequency \(\omega\), choose the centre-dependent gain
\(G(c)=\cos(\omega c)\). Thus
\[
S_\omega^{\cos}(x)
:=
\int_{\mathbb R}\cos(\omega c)B(x-c)\,dc.
\]
Changing variables \(t=x-c\), so \(c=x-t\), gives
\[
S_\omega^{\cos}(x)
=
\int_{\mathbb R}\cos(\omega(x-t))B(t)\,dt.
\]
Here \(x\) is fixed while integrating over \(c\), so \(dt=-dc\); the minus
sign is absorbed by reversing the integration limits.
Using
\[
\cos(\omega(x-t))
=
\cos(\omega x)\cos(\omega t)
+
\sin(\omega x)\sin(\omega t),
\]
we obtain
\[
S_\omega^{\cos}(x)
=
\cos(\omega x)\int_{\mathbb R}\cos(\omega t)B(t)\,dt
+
\sin(\omega x)\int_{\mathbb R}\sin(\omega t)B(t)\,dt.
\]
The second integral is zero because \(B(t)=\operatorname{sech}(\kappa t)\) is even and
\(\sin(\omega t)\) is odd, so the integrand is odd and is integrated over
the symmetric domain \(\mathbb R\). Therefore
\[
S_\omega^{\cos}(x)
=
C_\omega\cos(\omega x),
\qquad
C_\omega:=
\int_{\mathbb R}\cos(\omega t)B(t)\,dt.
\]
For the hyperbolic secant, the standard Fourier-transform identity gives the
cosine-transform formula\cite[Sec.~3.8]{kartner2005ultrafast}
\[
\int_{\mathbb R}\operatorname{sech}(t)\cos(qt)\,dt
=
\pi\operatorname{sech}\!\left(\frac{\pi q}{2}\right)
\qquad (q\in\mathbb R).
\]
Applying this with \(q=\omega/\kappa\) after the change of variable
\(t'=\kappa t\), we obtain
\[
C_\omega
=
\frac{\pi}{\kappa}
\operatorname{sech}\!\left(\frac{\pi\omega}{2\kappa}\right)
>0.
\]
Since \(C_\omega>0\), we can divide by \(C_\omega\):
\[
\cos(\omega x)
=
\frac{1}{C_\omega}
\int_{\mathbb R}\cos(\omega c)B(x-c)\,dc.
\]
So a cosine is an exact continuous superposition of translated bumps.

The sine case is similar: choosing \(G(c)=\sin(\omega c)\) gives
\[
\sin(\omega x)
=
\frac{1}{C_\omega}
\int_{\mathbb R}\sin(\omega c)B(x-c)\,dc.
\]

We move back to the discrete case of the filters by approximating the
continuous superposition over centres with a finite Riemann sum. Write
\(I=[x_{\min},x_{\max}]\). For \(\rho>0\), first keep only centres in the
finite interval \([x_{\min}-\rho,x_{\max}+\rho]\), so that
\[
\begin{aligned}
\int_{\mathbb R}\frac{\cos(\omega c)}{C_\omega}B(x-c)\,dc
&=
\int_{x_{\min}-\rho}^{x_{\max}+\rho}\frac{\cos(\omega c)}{C_\omega}B(x-c)\,dc\\
&\quad+
\int_{\mathbb R\setminus[x_{\min}-\rho,x_{\max}+\rho]}
\frac{\cos(\omega c)}{C_\omega}B(x-c)\,dc .
\end{aligned}
\]
If \(x\in I\) and \(c\notin[x_{\min}-\rho,x_{\max}+\rho]\), then \(|x-c|\ge\rho\). Since
\(B\) decays exponentially and \(|\cos(\omega c)/C_\omega|\le 1/C_\omega\),
the second integral tends to zero as \(\rho\to\infty\), uniformly for
\(x\in I\). On the remaining finite interval of centres, the integrand is
continuous on a compact set, so the first integral can be approximated
uniformly in \(x\in I\) by a Riemann sum, giving

\[
\sum_{k=1}^{K}
\frac{\Delta c\,\cos(\omega c_k)}{C_\omega}
B(x-c_k),
\]
and similarly for sine. Therefore finite signed sums of translated bumps can
approximate any sine or cosine on a compact interval. A trigonometric polynomial is defined as:

\[
T(x)
=
a_0+\sum_{m=1}^M
\left(a_m\cos(\omega_m x)+b_m\sin(\omega_m x)\right).
\]

By linearity, a single idealised PhyKAN edge with sufficiently many filters
can approximate any trigonometric polynomial. To see this, approximate each
sine and cosine term by its own finite filter sum, then collect all those
filters into one larger finite filter sum. This is still one element of
\(\mathcal F_{\rm edge}(I)\). Constants are included by the same argument with
\(\omega=0\). In that case \(C_0=\pi/\kappa>0\) and \(\cos(0x)=1\), so the
cosine construction approximates the constant function \(1\); scaling the
gains gives any constant \(a_0\).

\subsection{Trigonometric polynomials can approximate 1d functions}

We use the Stone--Weierstrass theorem in the form stated by
Pinkus\cite{pinkus2000weierstrass_supp}. In words, on a compact set, any family
of continuous real-valued functions that contains the constants, is closed
under addition, multiplication and scalar multiplication, and separates points
can uniformly approximate every continuous real-valued function on that set.

In our case the compact set is the interval \(I\subset\mathbb{R}\). The
approximating family is the set of trigonometric polynomials on \(I\), namely
finite sums of the form
\[
a_0+\sum_{m=1}^{M}
\left(a_m\cos(\omega_m x)+b_m\sin(\omega_m x)\right).
\]
These functions are continuous and contain the constants. They are also closed
under addition, scalar multiplication and multiplication; for multiplication,
products of sine and cosine terms reduce to finite sums of sine and cosine
terms by the standard product-to-sum identities.

It remains only to check that they separate points. If \(x\ne y\), choose
\[
\omega=\frac{\pi}{2(y-x)}.
\]
Then
\[
t\mapsto \sin(\omega(t-x))
\]
is a trigonometric polynomial, since it is a linear combination of
\(\sin(\omega t)\) and \(\cos(\omega t)\). Moreover it takes the value \(0\)
at \(t=x\) and the value \(1\) at \(t=y\). Thus the trigonometric polynomials
separate points.

Therefore all hypotheses of Stone--Weierstrass are satisfied, and
trigonometric polynomials uniformly approximate every continuous real-valued
function on \(I\).

\subsection{A single idealised PhyKAN edge is a one-dimensional universal approximator}

We now put together the two approximation results. Let \(I\subset\mathbb{R}\)
be a compact interval and let \(f:I\to\mathbb{R}\) be continuous. Let
\(\varepsilon>0\). From the Stone--Weierstrass argument in the previous
subsection, there is a trigonometric polynomial \(T\) such that
\[
\sup_{x\in I}|f(x)-T(x)|<\frac{\varepsilon}{2}.
\]

By the translated-bump approximation of trigonometric polynomials, finite
signed sums of translated bumps can approximate \(T\) on \(I\). Therefore the
same \(T\) can be approximated by a single idealised PhyKAN edge, say
\[
\mathcal E(x)=\sum_{\ell=1}^{L}G_\ell B(x-c_\ell),
\]
with
\[
\sup_{x\in I}|T(x)-\mathcal E(x)|<\frac{\varepsilon}{2}.
\]
Here the centres \(c_\ell\) are set by the filter corner frequencies, and the
coefficients \(G_\ell\) are the filter gains.

The triangle inequality then gives
\[
\sup_{x\in I}|f(x)-\mathcal E(x)|
\le
\sup_{x\in I}|f(x)-T(x)|
+
\sup_{x\in I}|T(x)-\mathcal E(x)|
<\varepsilon.
\]
Thus a single idealised PhyKAN edge can approximate any continuous
one-dimensional function on a compact interval.

\subsection{A PhyKAN network is a universal function approximator}

The preceding subsections were spent approximating a linear map, which a
conventional network obtains without cost. This is deliberate: the physical
device provides no linear maps ahead of its nonlinearities. It does, however,
provide fixed sigmoids at the nodes, and the argument below uses them.

Let $Q=I_1\times\cdots\times I_d\subset\mathbb{R}^d$ be compact and let
$F:Q\to\mathbb{R}$ be continuous. Let \(\varepsilon>0\). By the standard sigmoid universal
approximation theorem of Cybenko\cite{cybenko1989approximation}, there is a finite
one-hidden-layer sigmoid network
\[
N(x)=\sum_{j=1}^{n_{\mathrm h}}
\gamma_j\,\sigma_s(y_j(x)),
\qquad
y_j(x)=\sum_{i=1}^d r_{ij}x_i+b_j,
\]
such that
\[
\sup_{x\in Q}|F(x)-N(x)|<\frac{\varepsilon}{2}.
\]
Here \(\gamma_j\) are the usual output weights of the classical ANN, and
\[
\sigma_s(z)=\frac{1}{1+e^{-z/s_\sigma}}
\]
is the sigmoid used in the classical approximating network, with fixed scale
parameter \(s_\sigma>0\).

We now replace this classical ANN by a PhyKAN network. The PhyKAN architecture
already contains the same fixed, non-trainable sigmoid at hidden nodes. We replace the scalar weighted
sums entering the sigmoid by sums of one-dimensional PhyKAN edges, and then
replace the output weights by output PhyKAN edges.

For each hidden unit, split the affine preactivation \(y_j\) into
one-dimensional functions,
\[
g_{1j}(x_1)=r_{1j}x_1+b_j,\qquad
g_{ij}(x_i)=r_{ij}x_i\quad (i=2,\ldots,d),
\]
so that \(y_j(x)=\sum_i g_{ij}(x_i)\). The bias is assigned to the first
coordinate only to include it once; equivalently, it could be distributed
among the one-dimensional terms. By the one-dimensional universal
approximation result for idealised PhyKAN edges, each scalar function
\(g_{ij}:I_i\to\mathbb{R}\) can be uniformly approximated on \(I_i\). Fix an error
tolerance \(\eta>0\), to be chosen later, and for each pair \((i,j)\) choose
an idealised PhyKAN edge \(\mathcal E_{ij}:I_i\to\mathbb{R}\) such that
\[
\sup_{x_i\in I_i}|g_{ij}(x_i)-\mathcal E_{ij}(x_i)|<\eta .
\]
The classical preactivation \(y_j(x)=\sum_i g_{ij}(x_i)\) is then replaced
by the sum of these one-dimensional edge approximations:
\[
\widetilde y_j(x)=\sum_{i=1}^d \mathcal E_{ij}(x_i).
\]
Then, uniformly for \(x\in Q\),
\[
|y_j(x)-\widetilde y_j(x)|
\le
\sum_{i=1}^d |g_{ij}(x_i)-\mathcal E_{ij}(x_i)|
< d\eta .
\]

In the classical network, hidden unit \(j\) sends the scalar activation
\(\sigma_s(y_j(x))\) to the output weight \(\gamma_j\), which returns
\(\gamma_j\sigma_s(y_j(x))\). In the PhyKAN replacement, the corresponding
hidden activation is \(\sigma_s(\widetilde y_j(x))\), and the output weight
\(\gamma_j\) is replaced by a PhyKAN edge \(\mathcal O_j\).

Both \(\sigma_s(y_j(x))\) and \(\sigma_s(\widetilde y_j(x))\) lie in
\([0,1]\) because they are sigmoid outputs, so \(\mathcal O_j\) needs only
to approximate multiplication by \(\gamma_j\) on the interval \([0,1]\).
We therefore choose \(\mathcal O_j\) such that
\[
\sup_{u\in[0,1]}|\gamma_j u-\mathcal O_j(u)|<\eta,
\]
where \(u\) is a dummy variable ranging over possible scalar activation
values. Here \(\mathcal O_j\) is the output edge from hidden unit \(j\) to
the scalar output node. The output node sums these edge responses and does
not apply a hidden-node sigmoid.

The resulting PhyKAN network is
\[
F_{\rm P}(x)
=
\sum_{j=1}^{n_{\mathrm h}}
\mathcal O_j\!\left(
\sigma_s\!\left(\widetilde y_j(x)\right)
\right).
\]

It remains to show that this replacement is close to \(N\). The only
nonlinear point is that the sigmoid sees \(\widetilde y_j(x)\) instead of
\(y_j(x)\). This is controlled by the mean value theorem. Indeed,
\[
\sigma_s'(z)
=
\frac{1}{s_\sigma}\sigma_s(z)(1-\sigma_s(z)).
\]
Since \(0\le \sigma_s(z)\le 1\), we have
\(\sigma_s(z)(1-\sigma_s(z))\le 1/4\). Hence
\[
|\sigma_s'(z)|\le \frac{1}{4s_\sigma}=:L_\sigma .
\]
The factor \(1/s_\sigma\) appears because the exponent in \(\sigma_s\) is
\(-z/s_\sigma\). The mean value theorem therefore gives
\[
|\sigma_s(u)-\sigma_s(v)|\le L_\sigma |u-v|.
\]

We now compare one hidden-unit contribution in the classical ANN with its
PhyKAN replacement. The classical term is
\[
\gamma_j\sigma_s(y_j(x)),
\]
whereas the PhyKAN term is
\[
\mathcal O_j(\sigma_s(\widetilde y_j(x))).
\]
There are two errors: first, the sigmoid is fed the slightly wrong input
\(\widetilde y_j(x)\) instead of \(y_j(x)\); second, the output linear map
\(u\mapsto \gamma_j u\) is replaced by the output PhyKAN edge
\(\mathcal O_j\). To separate these two effects, insert the intermediate term
\(\gamma_j\sigma_s(\widetilde y_j(x))\). For each \(j\),
\[
\begin{aligned}
&
\left|
\gamma_j\sigma_s(y_j(x))-\mathcal O_j(\sigma_s(\widetilde y_j(x)))
\right|\\
&\le
\left|\gamma_j\sigma_s(y_j(x))-\gamma_j\sigma_s(\widetilde y_j(x))\right|
\\
&\quad+
\left|\gamma_j\sigma_s(\widetilde y_j(x))
-\mathcal O_j(\sigma_s(\widetilde y_j(x)))\right|\\
\end{aligned}
\]
To bound the first term, factor out \(\gamma_j\):
\[
\begin{aligned}
&
\left|\gamma_j\sigma_s(y_j(x))-\gamma_j\sigma_s(\widetilde y_j(x))\right|\\
&=
|\gamma_j|\,
\left|\sigma_s(y_j(x))-\sigma_s(\widetilde y_j(x))\right|\\
&\le
|\gamma_j|L_\sigma |y_j(x)-\widetilde y_j(x)|\\
&<
|\gamma_j|L_\sigma d\eta .
\end{aligned}
\]
Here the last inequality uses the preactivation bound
\(|y_j(x)-\widetilde y_j(x)|<d\eta\) proved above.

It remains to bound the second term
\[
\left|\gamma_j\sigma_s(\widetilde y_j(x))
-\mathcal O_j(\sigma_s(\widetilde y_j(x)))\right|.
\]
This term compares the two scalar maps \(u\mapsto\gamma_j u\) and
\(\mathcal O_j\) when both are evaluated at the same scalar input. That
common scalar input is
\[
\sigma_s(\widetilde y_j(x)),
\]
the sigmoid output of the replacement preactivation. Because the sigmoid
takes values in \([0,1]\), this input lies in \([0,1]\).

By construction, \(\mathcal O_j\) satisfies the uniform bound
\[
\sup_{u\in[0,1]}|\gamma_j u-\mathcal O_j(u)|<\eta,
\]
that is, \(|\gamma_j u-\mathcal O_j(u)|<\eta\) holds for every
\(u\in[0,1]\). Since \(\sigma_s(\widetilde y_j(x))\in[0,1]\), the bound
applies at this particular point and gives
\[
\left|\gamma_j\sigma_s(\widetilde y_j(x))
-\mathcal O_j(\sigma_s(\widetilde y_j(x)))\right|
<\eta.
\]
Therefore, for each \(j\),
\[
\left|
\gamma_j\sigma_s(y_j(x))-\mathcal O_j(\sigma_s(\widetilde y_j(x)))
\right|
<
|\gamma_j|L_\sigma d\eta+\eta .
\]

Now sum the hidden-unit errors. For each \(x\in Q\),
\[
\begin{aligned}
N(x)-F_{\rm P}(x)
&=
\sum_{j=1}^{n_{\mathrm h}}
\left[
\gamma_j\sigma_s(y_j(x))
-
\mathcal O_j(\sigma_s(\widetilde y_j(x)))
\right],
\end{aligned}
\]
so the triangle inequality gives
\[
\begin{aligned}
|N(x)-F_{\rm P}(x)|
&=
\left|
\sum_{j=1}^{n_{\mathrm h}}
\left[
\gamma_j\sigma_s(y_j(x))
-
\mathcal O_j(\sigma_s(\widetilde y_j(x)))
\right]
\right|\\
&\le
\sum_{j=1}^{n_{\mathrm h}}
\left|
\gamma_j\sigma_s(y_j(x))
-
\mathcal O_j(\sigma_s(\widetilde y_j(x)))
\right|\\
&<
\sum_{j=1}^{n_{\mathrm h}}
\left(|\gamma_j|L_\sigma d\eta+\eta\right)\\
&=
\left(n_{\mathrm h}
+dL_\sigma\sum_{j=1}^{n_{\mathrm h}}|\gamma_j|\right)\eta .
\end{aligned}
\]
The inequality above holds for every input vector
\(x=(x_1,\ldots,x_d)\in Q=I_1\times\cdots\times I_d\). Therefore the same
right-hand side also bounds the supremum of the error over the whole input
domain:
\[
\sup_{x\in Q}|N(x)-F_{\rm P}(x)|
\le
\left(n_{\mathrm h}+dL_\sigma\sum_{j=1}^{n_{\mathrm h}}|\gamma_j|\right)\eta .
\]
Because the factor multiplying \(\eta\) is finite, choose \(\eta\) small
enough that
\[
\sup_{x\in Q}|N(x)-F_{\rm P}(x)|<\frac{\varepsilon}{2}.
\]
Finally, the triangle inequality gives
\[
\sup_{x\in Q}|F(x)-F_{\rm P}(x)|
\le
\sup_{x\in Q}|F(x)-N(x)|
+
\sup_{x\in Q}|N(x)-F_{\rm P}(x)|
<\varepsilon .
\]
Thus PhyKAN networks are universal approximators on compact domains.

\begin{remark}
The statement is an idealised density result: the number of filters and the
hidden width may grow, centres and gains are continuous, and finite hardware
range, quantisation, and transfer mismatch contribute implementation error.
\end{remark}

\section{Benchmarking Experimental Transfer}
To quantify model-to-hardware transfer quality in a task-agnostic manner, we evaluate the mean-squared error between simulated and experimental edges across transferred networks in the kinematics and reinforcement-learning control tasks.

\begin{figure}[!htbp]
    \centering
    \includegraphics[width=0.75\linewidth]{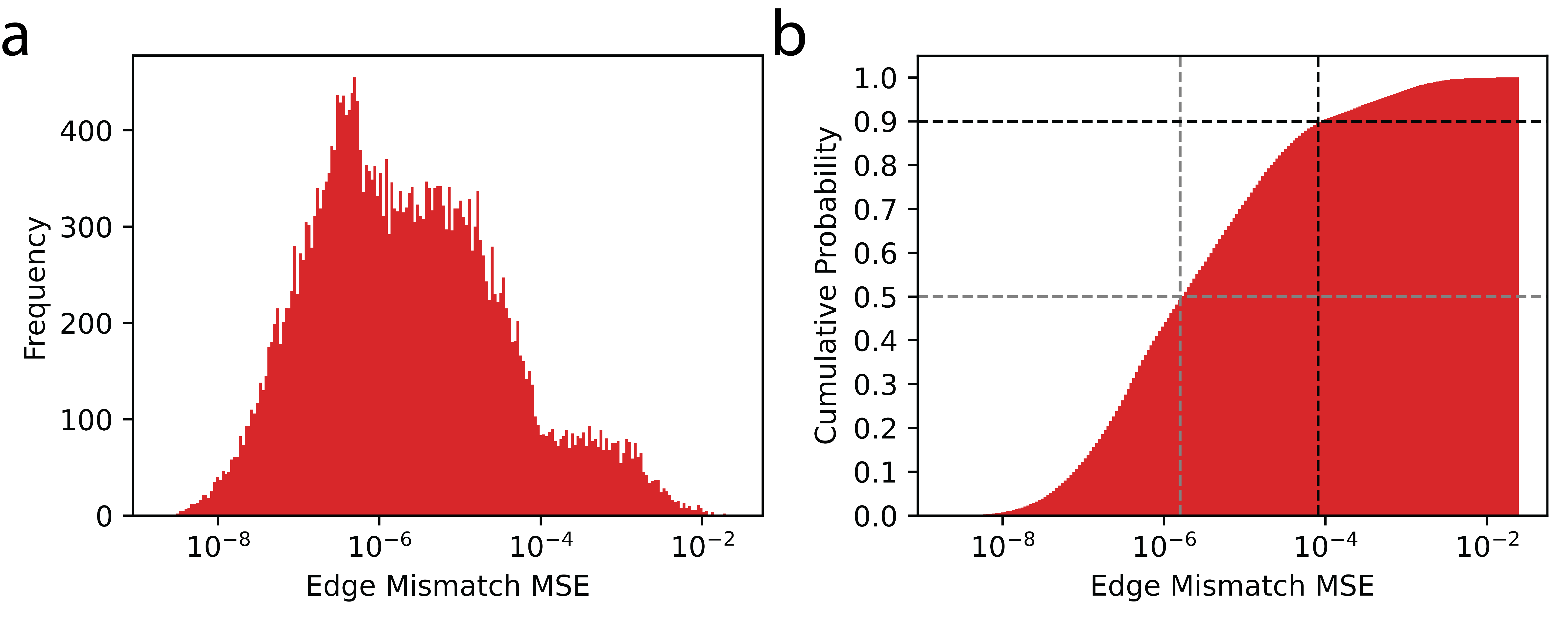}
    \caption{\textbf{Transfer Error Statistics}- (a) Histogram of mean-squared errors between simulated edge response and experimentally realised edge response. (b) Cumulative distribution function of the same errors. The black dashed line marks the 90th percentile, at an MSE of \(7.92\times10^{-5}\), while the grey dashed line marks the median, at an MSE of \(1.58\times10^{-6}\).}
    \label{fig:transfer_errors}
\end{figure}

To reduce the need for additional experimentation, the trained networks used in Figs.~\ref{fig:Robotics}, \ref{fig:CartPole}, and \ref{fig:PV_Control} were used as representative samples of the range of edges likely to be realised in practice. Evaluation was performed by linearly spacing 200 inputs over the maximum input range (logarithmically spaced in frequency) for each edge, calculating the simulated response via the analytical equations introduced in Section~\ref{sec:transfunc}, and comparing the results to the experimentally gathered data. Mean-squared errors were then calculated for each of the edges, resulting in $\approx 35{,}000$ samples of edge transfer functions.
Supplementary Fig.~\ref{fig:transfer_errors}a shows a histogram of the mean-squared errors between simulation results and experimental data, while panel b shows the cumulative distribution function for the same data. Half of the edges have an MSE at or below $1.58\times10^{-6}$, and 90\% have an MSE at or below $7.92\times10^{-5}$.

\FloatBarrier

\section{Benchmarking on Classification Problems}
As well as the continuous-valued kinematics and control problems studied in the main text, the PhyKANs were also benchmarked on the standard classification benchmark of Fashion-MNIST. This task resembles a more difficult version of the classic MNIST digit recognition problem, and requires classifiers to assign input images to one of 10 clothing classes. Models here were trained for 100,000 iterations with a batch size of 50.

\begin{figure}[!htbp]
    \centering
    \includegraphics[width=0.75\linewidth]{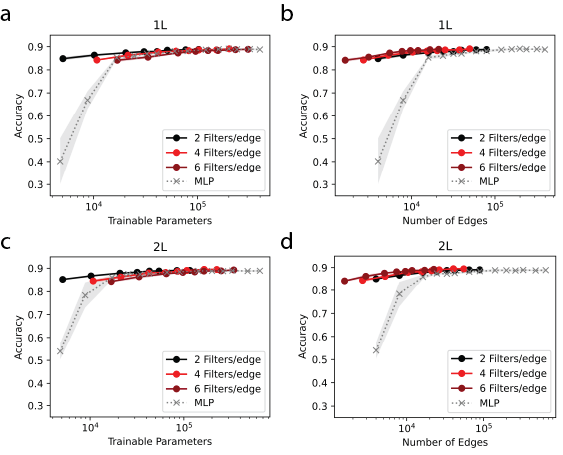}
    \caption{\textbf{Fashion-MNIST Performance.} Panels (a) and (c) show classification accuracy versus trainable parameters for networks with one and two hidden layers respectively, while panels (b) and (d) show the same accuracy data as a function of total network edges, again for one and two hidden layer networks. In all panels, data markers represent mean performance, while the shaded regions show the standard deviation of performance across 10 repetitions with different initialisations of networks.}
    \label{fig:FMNIST}
\end{figure}

Supplementary Fig.~\ref{fig:FMNIST} shows the final performance of simulated PhyKAN models on Fashion-MNIST as a function of the number of filters per edge, with an MLP used to provide baseline performance. Panels (a) and (c) plot the performance of PhyKANs with 2, 4, and 6 parallel filters per edge with respect to the total number of trainable parameters for networks with one and two hidden layers respectively. For this task, the number of filters per edge, which serves as a proxy for edge complexity, has limited effect on the overall performance of the network, with only smaller networks with fewer filters per edge showing improved per-parameter performance compared to MLP controls.
However, when edge complexity is increased, although the number of trainable parameters increases, the number of edges required for performance decreases because edges can be pruned more effectively from networks with more complex edge functions, as shown in panels (b) and (d). The PhyKANs also show a marked increase in performance compared to MLPs when normalised for total number of edges. Simpler connectivity topologies are an advantage for fabrication, since fewer physical connections must be routed and calibrated.

\FloatBarrier

\section{Benchmarking on Discrete Control}
The discrete CartPole control task environment follows the classic pole-balancing problem\cite{barto1983cartpole} and was taken directly from the OpenAI Gym library\cite{brockman2016openai}. The agent receives the state of the cart at every timestep $t$, $S_t$, defined by Eq.~\ref{eq:states_discrete}:
\begin{equation}
    S_t= [x_t, \dot{x}_t, \theta_t, \dot{\theta}_t]
    \label{eq:states_discrete}
\end{equation}
where $x_t$ denotes the position of the cart with respect to the centre of the environment at time $t$ in arbitrary units, $\dot{x}_t$ the velocity of the cart at time $t$ in arbitrary units per second, $\theta_t$ the angle of the pole with respect to the normal of the cart at time $t$ in radians, and $\dot{\theta}_t$ the angular velocity of the pole at time $t$ in radians per second. The agent chooses one of two discrete actions, moving the cart left or right, and is presented with a reward of +1 for every timestep for which it keeps the pole upright.
Rewards were shaped according to the same function as the continuous task. Training of the agent was performed using deep Q-learning with a separate target network\cite{mnih2015human_supp}. This approach features two networks (typically MLPs, though here also PhyKANs), conditioned on two sets of parameters $\bm{\theta}^0$ and $\bm{\theta}'$, which are responsible for predicting the total future returns of the current state--action pair, $Q(S_t, A_t;\bm{\theta}^0)$, and of the next state, $Q(S_{t+1}, a;\bm{\theta}')$, respectively. Parameters for the current network, $\bm{\theta}^0$, are updated via gradient descent, by minimising the following objective function defined in Eq.~\ref{eq:rl_objective}:
\begin{equation}
   \mathcal{L} = [R_{t+1}^s+\gamma\underset{a}{\mathrm{max}}Q(S_{t+1},a;\bm{\theta}')-Q(S_t, A_t;\bm{\theta}^0)]^2
   \label{eq:rl_objective}
\end{equation}
where $\gamma$ represents the discount factor used to discount future rewards, here selected as 0.95. Note that the maximising action in the target is both selected and evaluated by $\bm{\theta}'$; we did not use the decoupled action selection of double Q-learning\cite{vanhasselt2016double}.
The parameters of the network used for estimating total future returns of the next timestep, $\bm{\theta}'$, received soft updates such that parameters slowly converge to those of the current prediction network, $\bm{\theta}^0$, via the update rule shown in Eq.~\ref{eq:softupdate}:
\begin{equation}
    \bm{\theta}' \leftarrow \tau \bm{\theta}^0 + (1-\tau)\bm{\theta}'
    \label{eq:softupdate}
\end{equation}
where $\tau$ denotes the rate at which original parameters are overwritten, here 0.001.

Networks were optimised over 1000 episodes, with the actions selected according to a softmax policy with a temperature of 1 to encourage exploration. Performance was evaluated after each episode by repeating the run with a greedy policy, and no parameter updates were made during this evaluation.

\begin{figure}[!htbp]
    \centering
    \includegraphics[width=0.7\linewidth]{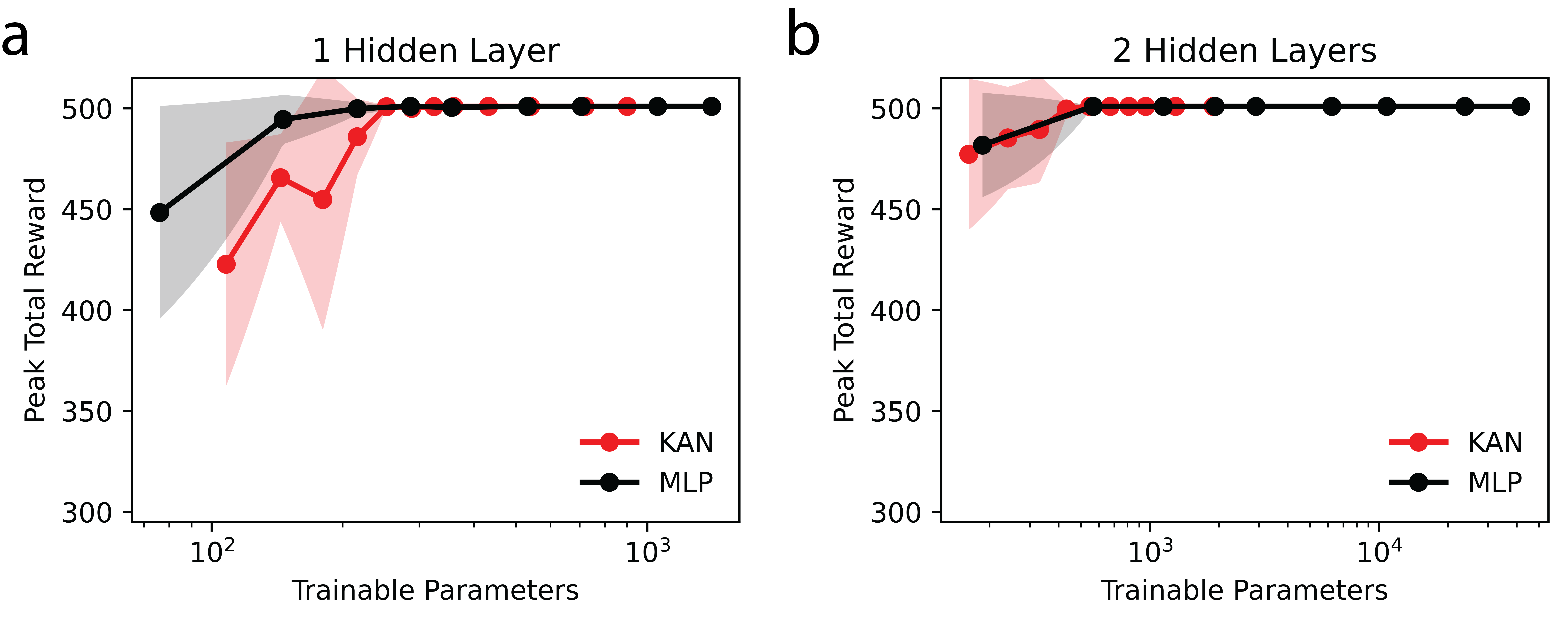}
    \caption{\textbf{CartPole Performance for Binary Action Selection}- Comparison of the peak performance of models trained using deep Q-learning with a target network, for networks with a single hidden layer (a) and two hidden layers (b). Marked data points show mean performance over 10 independent initialisations, while the shaded regions show the standard deviation.}
    \label{fig:binary_CP}
\end{figure}

Supplementary Fig.~\ref{fig:binary_CP} shows a performance comparison between simulated PhyKAN models and MLPs.

Both model classes are used in the deep Q-learning algorithm described above for the standard CartPole task, with either one or two hidden layers.

With a single hidden layer, both architectures reach maximum performance at around the same number of trainable parameters, though the MLP outperforms the PhyKAN for the smallest network sizes. In the two hidden layer experiments, performance between the two architectures is almost identical.

In the classical implementation of the CartPole benchmark for reinforcement learning, the agent receives the same continuous input data, but during action selection, the output is binarised into applying forces of $\pm10$ N. With greedy action selection, this approach resembles a winner-take-all algorithm operating on Q-values predicted by the network, making the task functionally similar to binary classification. We hypothesise that this explains the similarity to the Fashion-MNIST results and the reduction in parameter-efficiency gains compared with the continuous-valued kinematics and control tasks in the main manuscript.

\section{Intrinsic Dimensionality Analysis of Learned Representations}\label{sec:dimensionality}
\begin{figure}[!htbp]
    \includegraphics[width=1\linewidth]{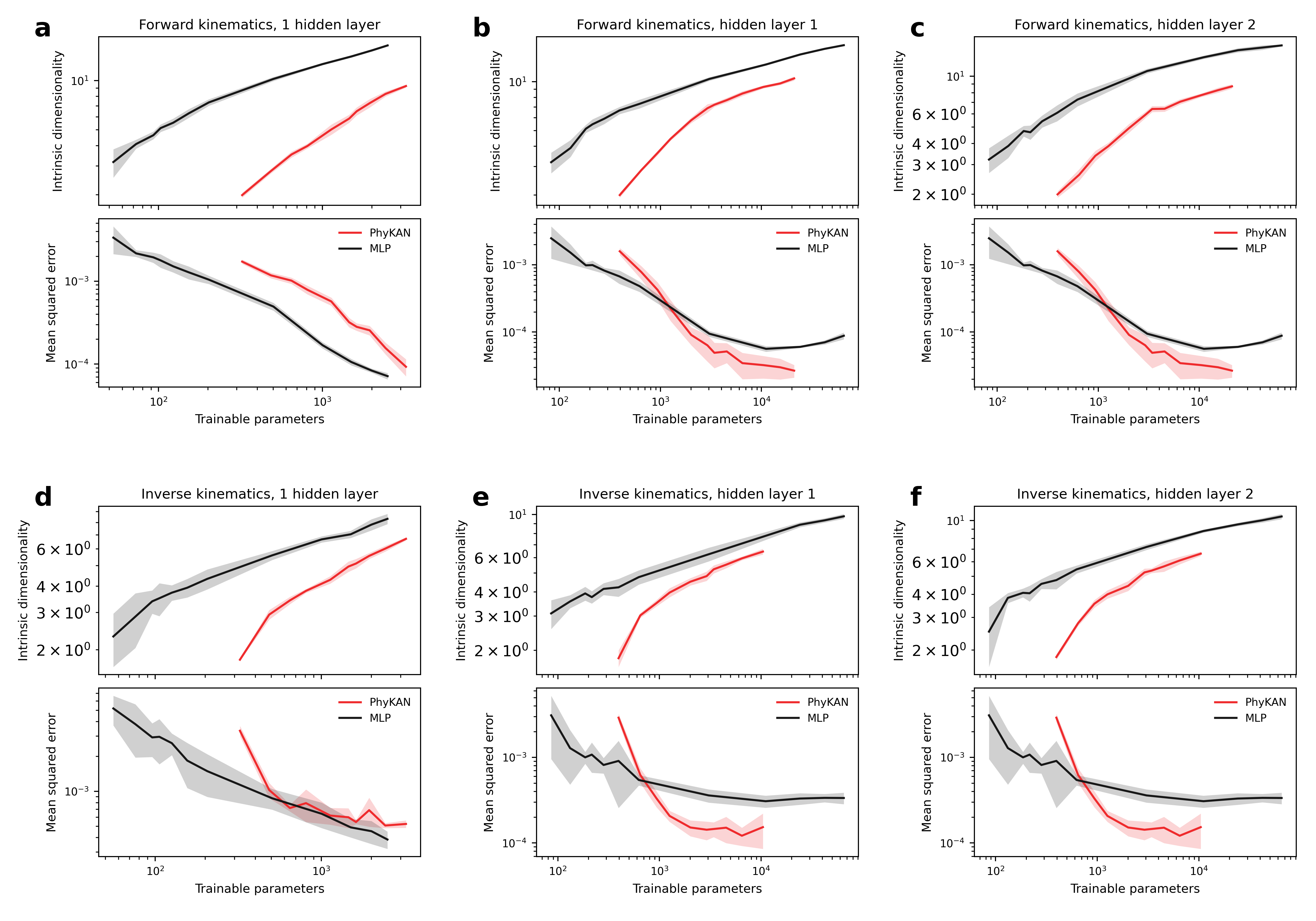}
    \caption{\textbf{Intrinsic Dimensionality Analysis of PhyKAN and MLP for Six-Axis Robot Kinematics.} Separability-based intrinsic dimensionality of centred and whitened post-activation hidden representations (upper subplots) and test mean-squared error (lower subplots) for forward (\textbf{a--c}) and inverse (\textbf{d--f}) kinematics. Panels \textbf{a,d} show one-hidden-layer networks; \textbf{b,e} show the first hidden layer of two-hidden-layer networks; and \textbf{c,f} show the second hidden layer. Curves are arithmetic means across 10 independently trained models and shaded regions show one population standard deviation. PhyKANs are red and MLPs are black.}
    \label{fig:dimensionality}
\end{figure}

To characterise the representations learned by the PhyKANs and MLPs, we quantify the intrinsic dimensionality of their hidden-layer representations using the metric of Sutton et al.\cite{sutton2023intrindim_supp}. According to this metric, data sampled from a distribution $\mathcal{D}$ on $\mathbb{R}^d$ have an intrinsic dimensionality \(n(\mathcal{D})\in\mathbb{R}\) with respect to a centre \(c\in\mathbb{R}^d\) if:
\begin{equation}
    P(x,y\sim\mathcal{D}:\langle x-y, y-c\rangle\geq0)=\frac{1}{2^{n(\mathcal{D})+1}}
    \label{intrinsic_dim}
\end{equation}
For each trained model and hidden layer, we centred and whitened the post-activation representation before applying the metric. The PhyKAN representation is the post-sigmoid output, including the learned threshold when enabled, and the MLP representation is the post-ReLU output. Supplementary Fig.~\ref{fig:dimensionality} compares these estimates as a function of trainable parameters for the six-axis kinematics models used in Fig.~\ref{fig:Robotics}.

Across all six panels, the estimated dimensionality generally increases with model size. At overlapping parameter values, the PhyKAN estimate is lower than the MLP estimate for the one-hidden-layer networks and for both hidden layers of the two-hidden-layer networks. The test-error curves show that estimated dimensionality can continue to increase while changes in mean test error are small. Generally, the KANs can find a solution in lower dimensional space without compromising accuracy, suggesting that they smootly model the noisy data.

\begin{figure}
    \centering
    \includegraphics[width=0.85\linewidth]{Kinematics_vs_edges.png}
    \caption{\textbf{Comparison of edges vs mean-squared-error for noised robotic kinematics tasks.} Mean-squared-error versus number of edges in a network for PhyKAN and MLPs for (a) 1-Layer forward kinematics, (b) 2-layer forward kinematics, (c) 1-layer inverse kinematics, and (d) 2-layer inverse kinematics.}
    \label{fig:robotics_vs_edgecount}
\end{figure}
\FloatBarrier

\begin{figure}
\centering
    \includegraphics[width=0.85\linewidth]{Noiseless_kinematics.png}
    \caption{\textbf{Effects of noisy data for hardware-discretised PhyKANs and unconstrained (continuous) PhyKANs.} Mean-squared error versus trainable parameter count for two-hidden-layer forward kinematics with exact data (a) and imperfect actuation uniformly distributed over $\pm0.1$ radians on each joint (b), as well as two-hidden-layer inverse kinematics for exact (c) and noisy (d) data. Red lines show unconstrained PhyKANs, brown lines hardware-discretised PhyKANs, and black lines MLPs. At larger network sizes, unconstrained PhyKANs achieve lower errors than MLPs in both tasks. Hardware discretisation reduces this advantage in forward kinematics, while it is retained in the reported inverse-kinematics comparison.}
    \label{fig:noiseless_robotics}
\end{figure}
\FloatBarrier
\section{L1 + Shannon Entropy Regularisation on MLPs}\label{sec:MLP_regularisation}
For a fair comparison with the regularised PhyKAN models, we applied analogous L1 and entropy penalties to the MLP baselines. The hardware-aware quantisation penalty was not applied to MLPs, since these models were not transferred to physical hardware. Standard L1 regularisation was performed on the weights of the MLP. The entropy penalty was computed on node activations in MLPs, not on the edge-wise activations described in Section~\ref{sec:regularisation}:
\begin{equation}
\mathcal{L}_{\mathrm{reg}}=\lambda\sum_\ell\left(\|\bm{\theta}^{(\ell)}\|_1-\sum_hq_h^{(\ell)}\log{q_h^{(\ell)}}\right)
\end{equation}
\begin{equation}
    q_h = \frac{|a_h|}{\sum_h |a_h|}.
\end{equation}
where $\lambda$ is the regularisation coefficient, $\bm{\theta}$ describes the model's weights, and $a_h$ is the activation on node $h$ in a given layer.

\begin{figure}
    \centering
    \includegraphics[width=\linewidth]{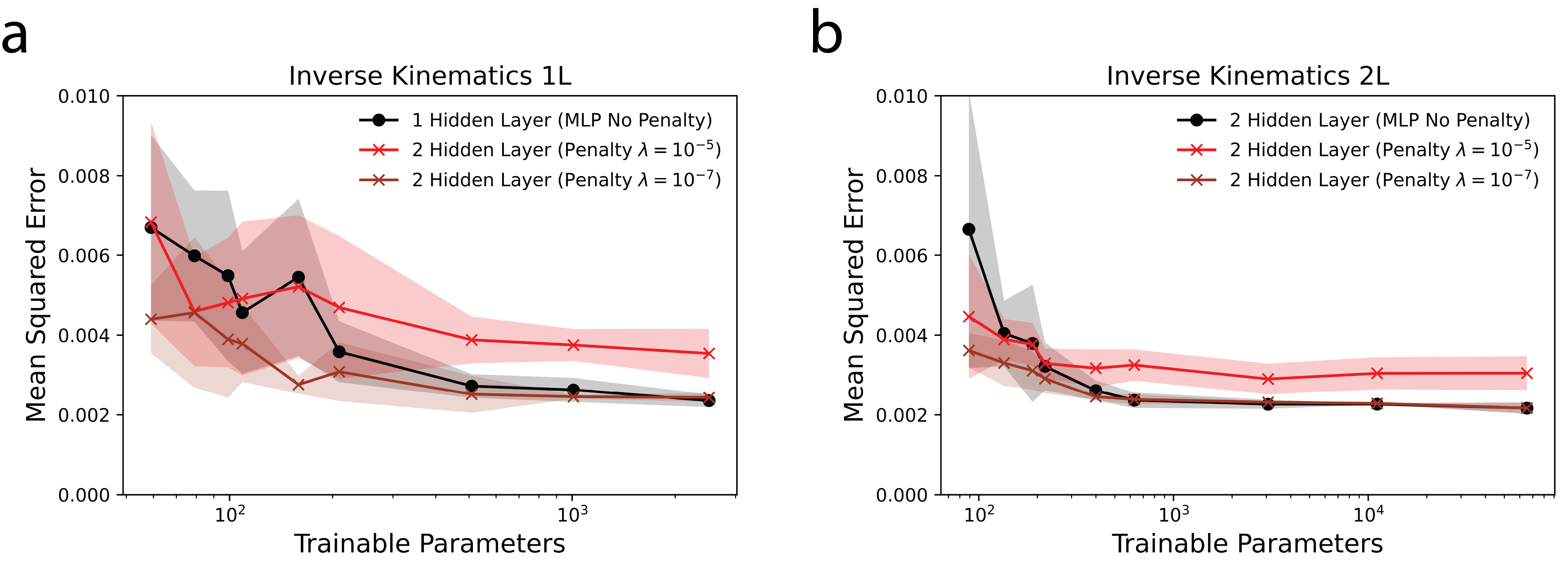}
    \caption{\textbf{Effect of regularisation on MLPs.} Comparison between mean-squared errors achieved on multilayer perceptrons with no regularisation (black), identical regularisation magnitude to the KANs (red), and a regularisation magnitude optimised for MLPs (brown) for the six-axis inverse kinematics problem. Panels (a) and (b) show errors achieved in networks with a single hidden layer, and two hidden layers respectively. Markers show the average performance across ten independent runs, while the shaded region reflects the standard deviation across the ten runs.}
    \label{fig:penalties}
\end{figure}

Supplementary Fig.~\ref{fig:penalties} shows a comparison of the mean-squared errors achieved on the inverse kinematics task for MLPs without regularisation (black), with the same regularisation magnitude of $1\times10^{-5}$ as the PhyKANs (red), and an optimised regularisation magnitude of $1\times10^{-7}$ (crimson). While regularisation boosts performance for the smallest MLPs in both cases compared to the control, in larger networks only the optimised magnitude leaves performance unaffected, whereas the stronger penalty reduces it. This suggests that in MLPs, regularisation can help find better solutions in constrained networks, but can form additional constraints that reduce accuracy if it is too strong.

\section{Hyperparameter optimisation of MLP models of Memristor circuits.}\label{sec:hyperparamMLP}
\begin{figure}
    \centering
    \includegraphics[width=\linewidth]{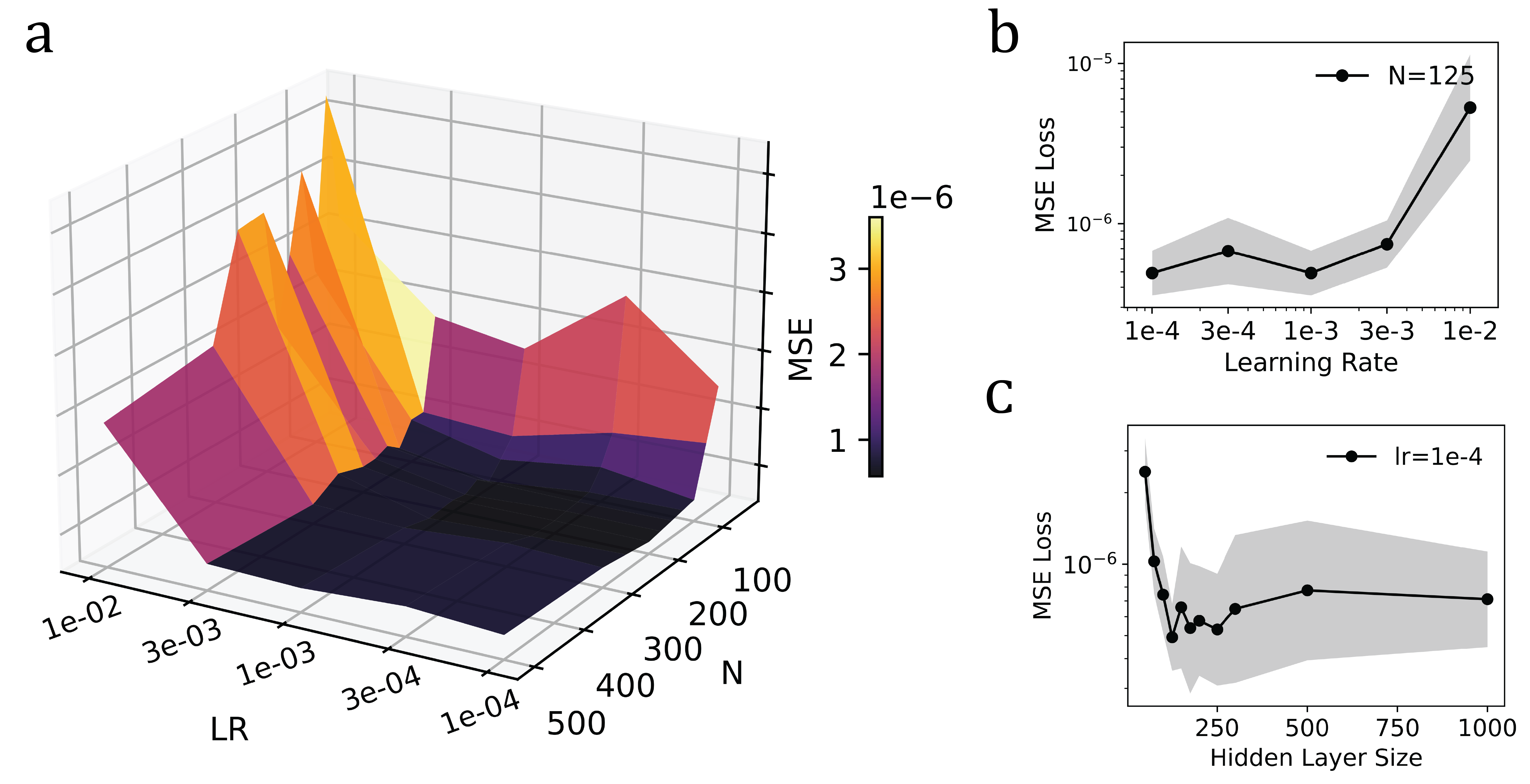}
    \caption{\textbf{Hyperparameter optimisation of memristor models.} (a) Surface plot of validation mean-squared error of the MLP model of the memristor circuit as a function of the number of nodes in the hidden layers and the learning rate used to train the network. Panels (b) and (c) show the lowest validation mean-squared errors as a function of learning rate and hidden-layer size, respectively.}
    \label{fig:memkanmlp}
\end{figure}

Hyperparameters of learning rate and number of nodes in the two hidden layers for the MLPs used to model the memristor circuits were selected using the lowest validation mean-squared error averaged across 10 independent runs on different shuffles of training and validation splits of size $3\times10^6$ and $5\times10^5$, respectively. Each model was trained for one million iterations with a fixed minibatch size of 10,000. Supplementary Fig.~\ref{fig:memkanmlp}(a) shows validation mean-squared error as a function of learning rate and number of neurons in each hidden layer, with lower validation error for more than 100 nodes and learning rates below $3\times 10^{-3}$. The best-performing combination was 125 nodes and a learning rate of $1\times10^{-4}$. Panels (b) and (c) show slices through this surface for validation mean-squared error versus learning rate and hidden-layer size, respectively.

\FloatBarrier

\section{Supporting Figures for Memristor-based PhyKANs}
\begin{figure}[!htbp]
\centering
\begin{subfigure}[t]{0.15\textwidth}
    \centering
    \includegraphics[width=\linewidth]{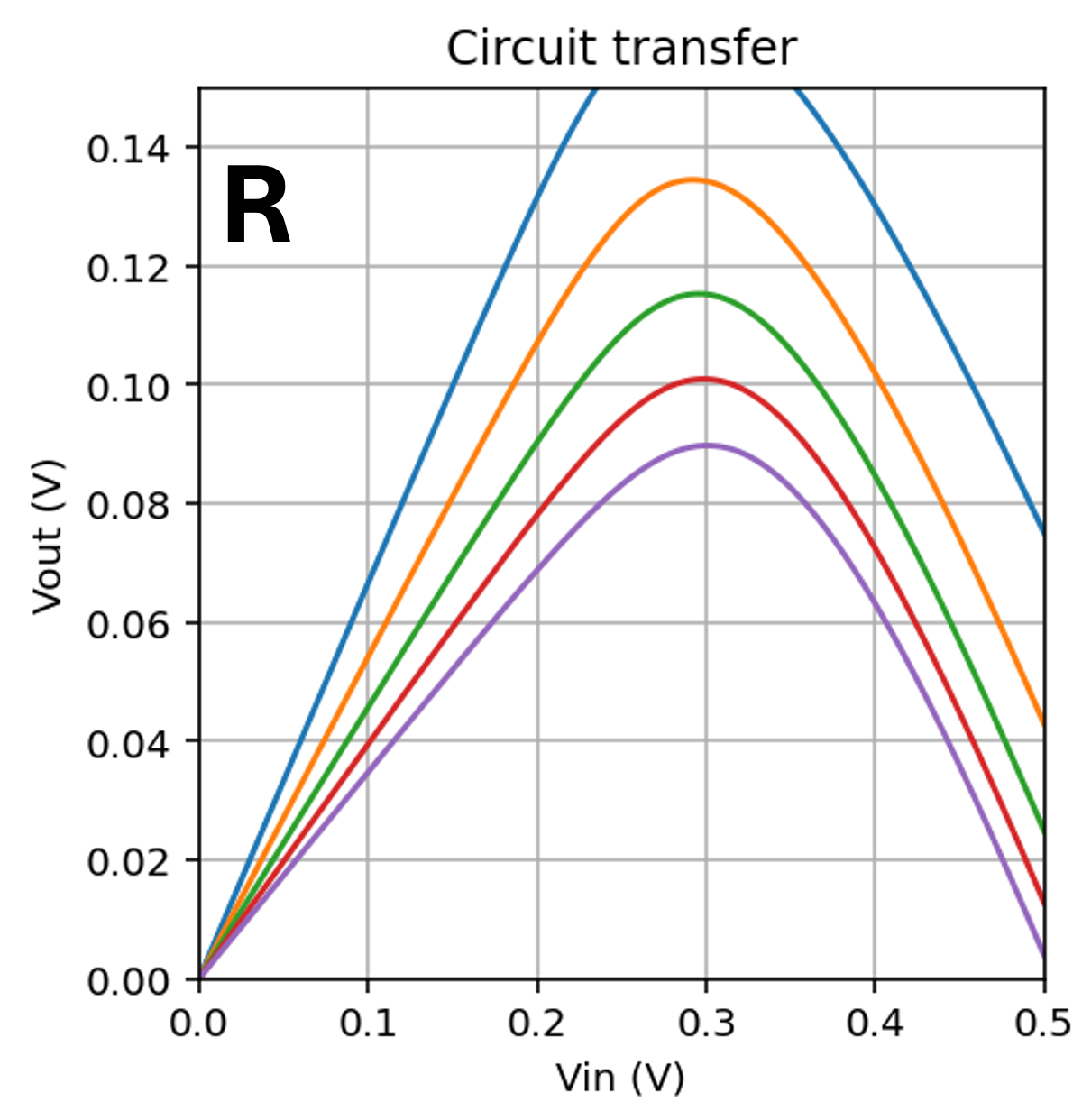}
    \caption{}
    \label{fig:R_BP_impact}
\end{subfigure}
\hfill
\begin{subfigure}[t]{0.15\textwidth}
    \centering
    \includegraphics[width=\linewidth]{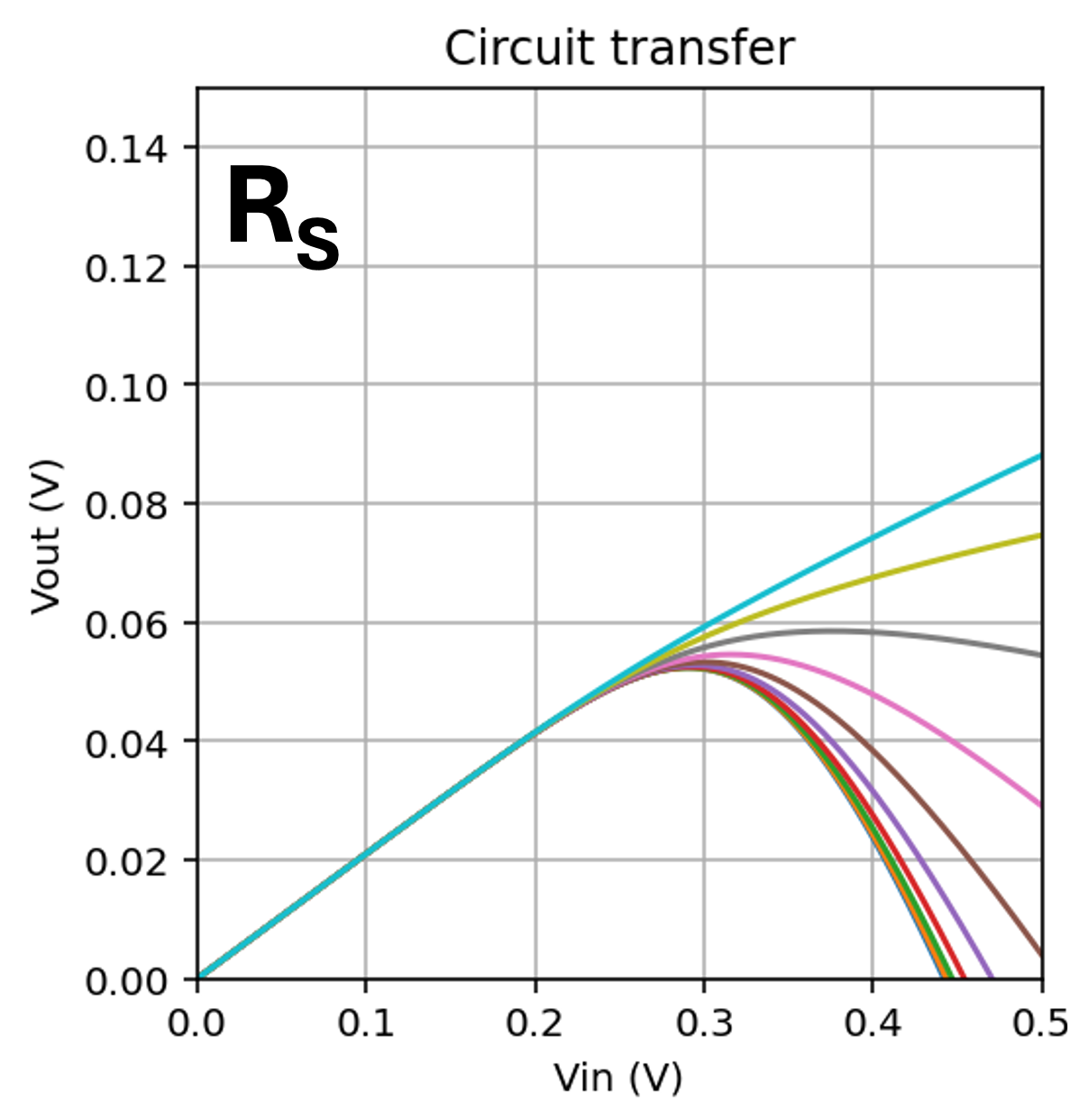}
    \caption{}
    \label{fig:RS_BP_impact}
\end{subfigure}
\hfill
\begin{subfigure}[t]{0.15\textwidth}
    \centering
    \includegraphics[width=\linewidth]{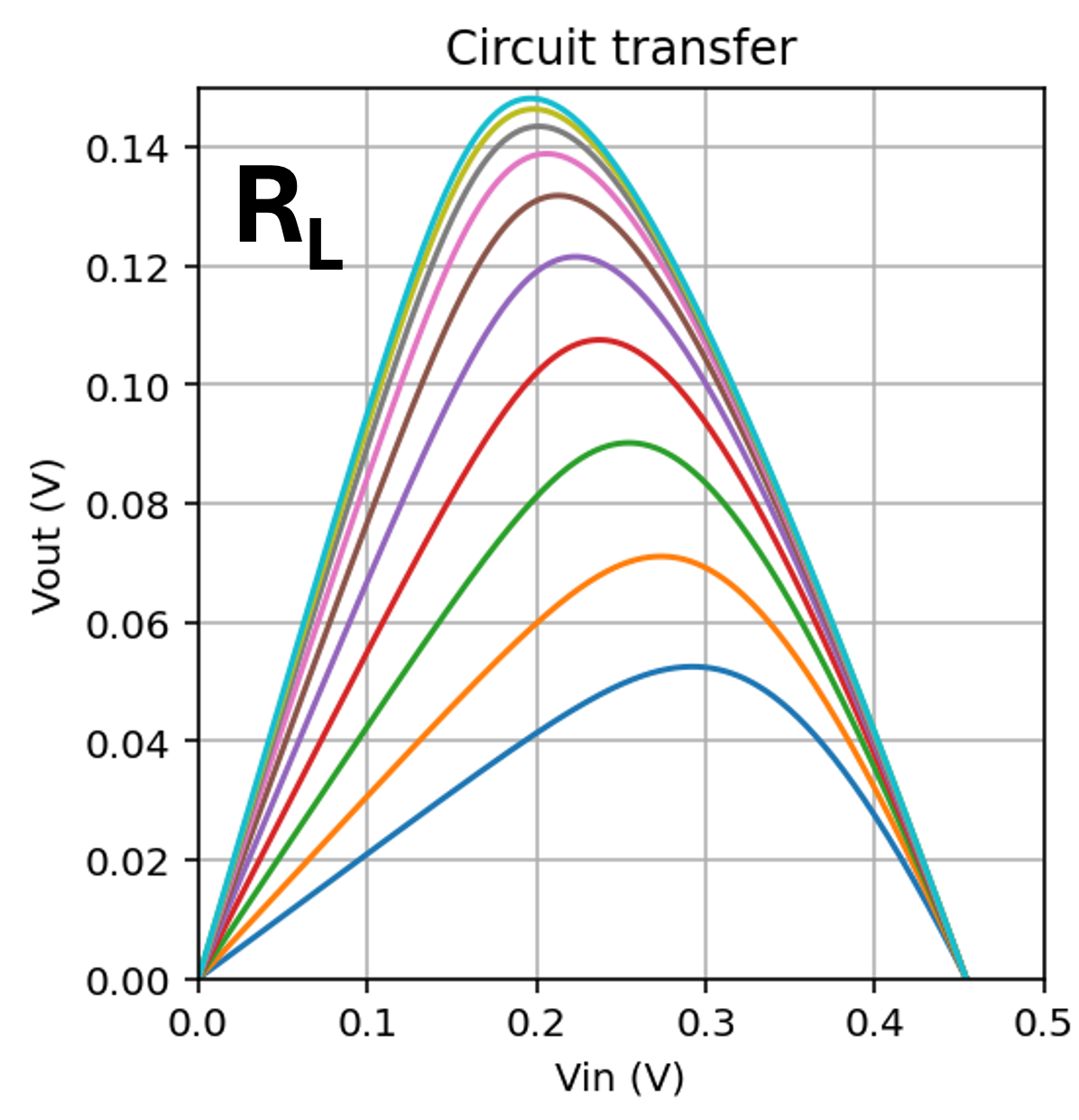}
    \caption{}
    \label{fig:RL_BP_impact}
\end{subfigure}
\hfill
\begin{subfigure}[t]{0.15\textwidth}
    \centering
    \includegraphics[width=\linewidth]{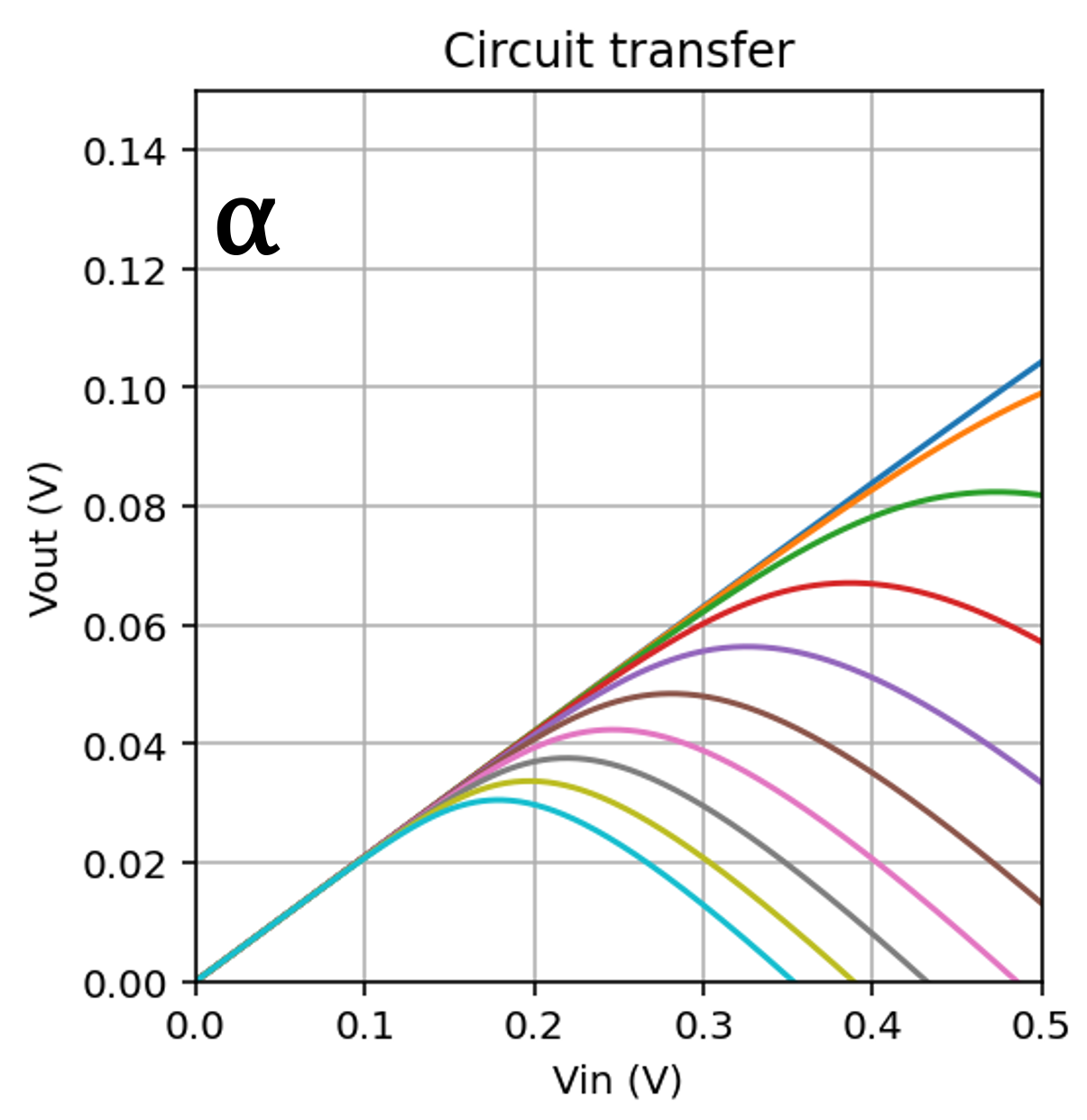}
    \caption{}
    \label{fig:alpha_BP_impact}
\end{subfigure}
\hfill
\begin{subfigure}[t]{0.15\textwidth}
    \centering
    \includegraphics[width=\linewidth]{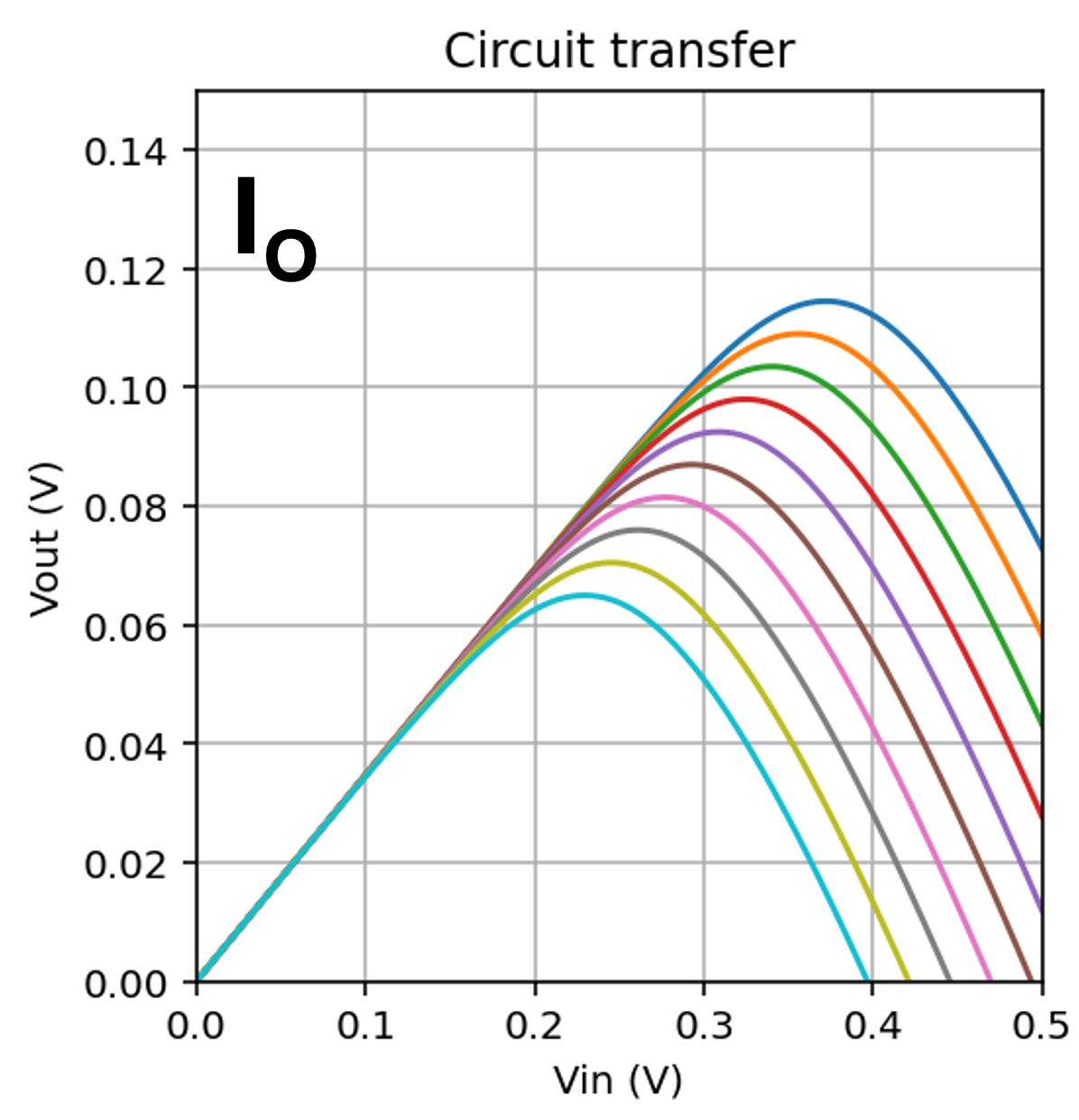}
    \caption{}
    \label{fig:Io_BP_impact}
\end{subfigure}

\caption{The specific transfer function of the memristor circuit is governed by five fitting parameters, whose impact in the band-pass case is indicated in panels (a)--(e).}
\label{fig:top_three_bottom}
\end{figure}
\begin{figure}[!htbp]
\centering

\begin{subfigure}[c]{0.48\textwidth}
    \centering
    \includegraphics[width=\linewidth]{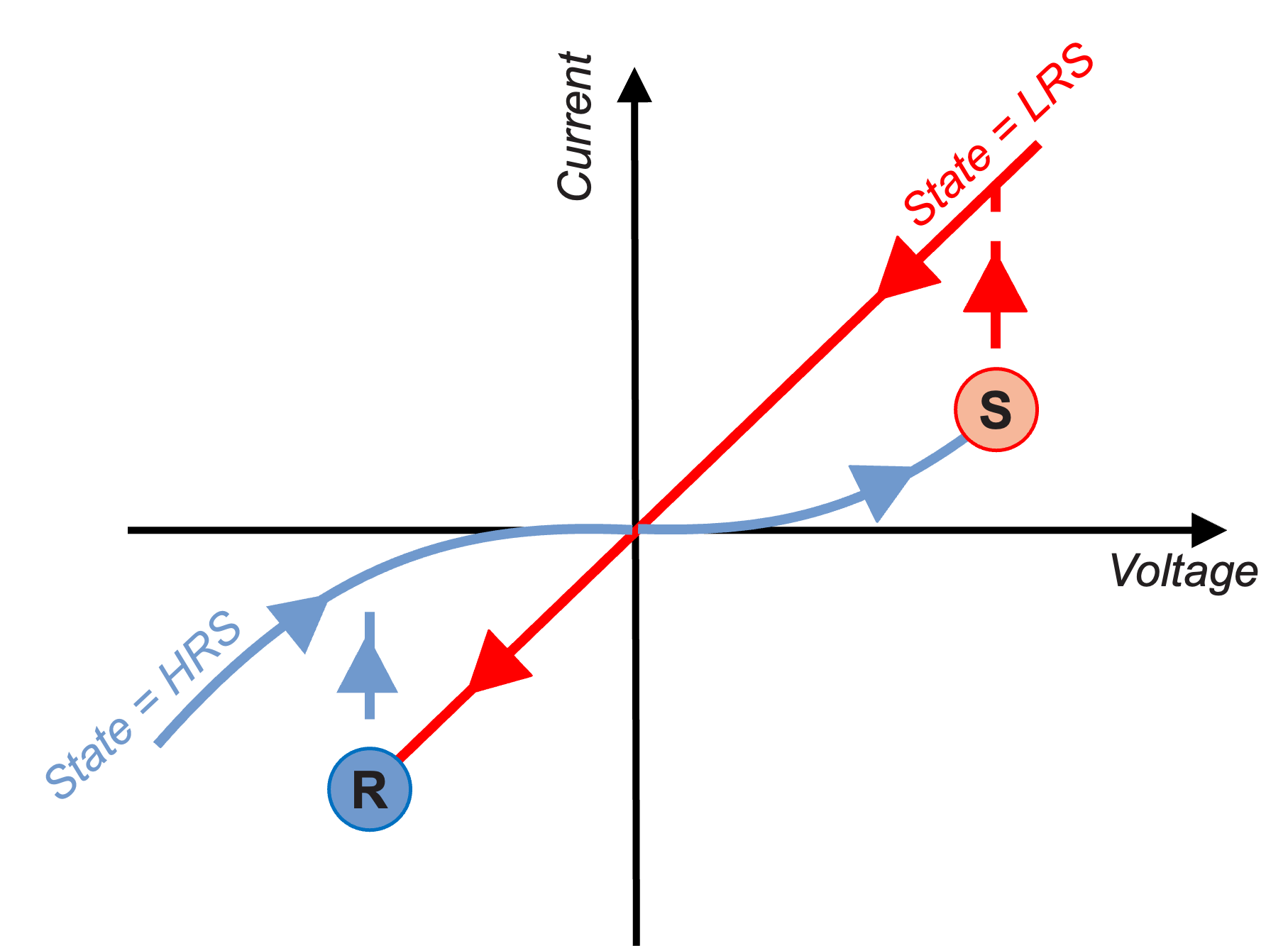}
    \caption{}
    \label{fig:memristor_IV}
\end{subfigure}
\hfill
\begin{minipage}[c]{0.48\textwidth}
    \centering

    \begin{subfigure}[t]{\linewidth}
        \centering
        \includegraphics[width=\linewidth]{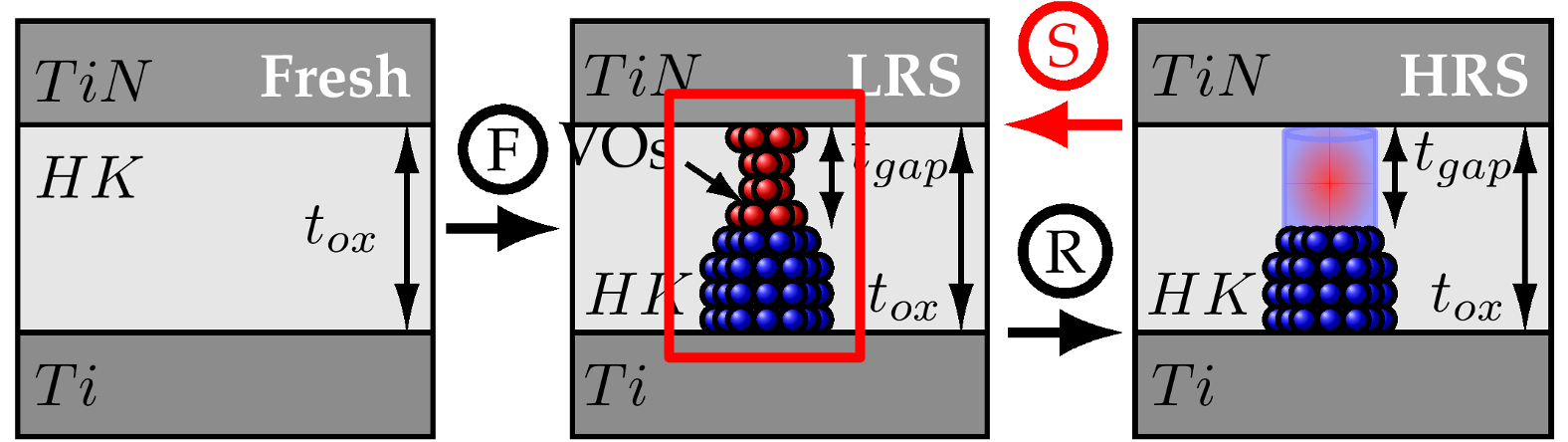}
        \caption{}
        \label{fig:memristor_physics}
    \end{subfigure}

    \vspace{0.5em}

    \begin{subfigure}[t]{\linewidth}
        \centering
        \includegraphics[width=\linewidth]{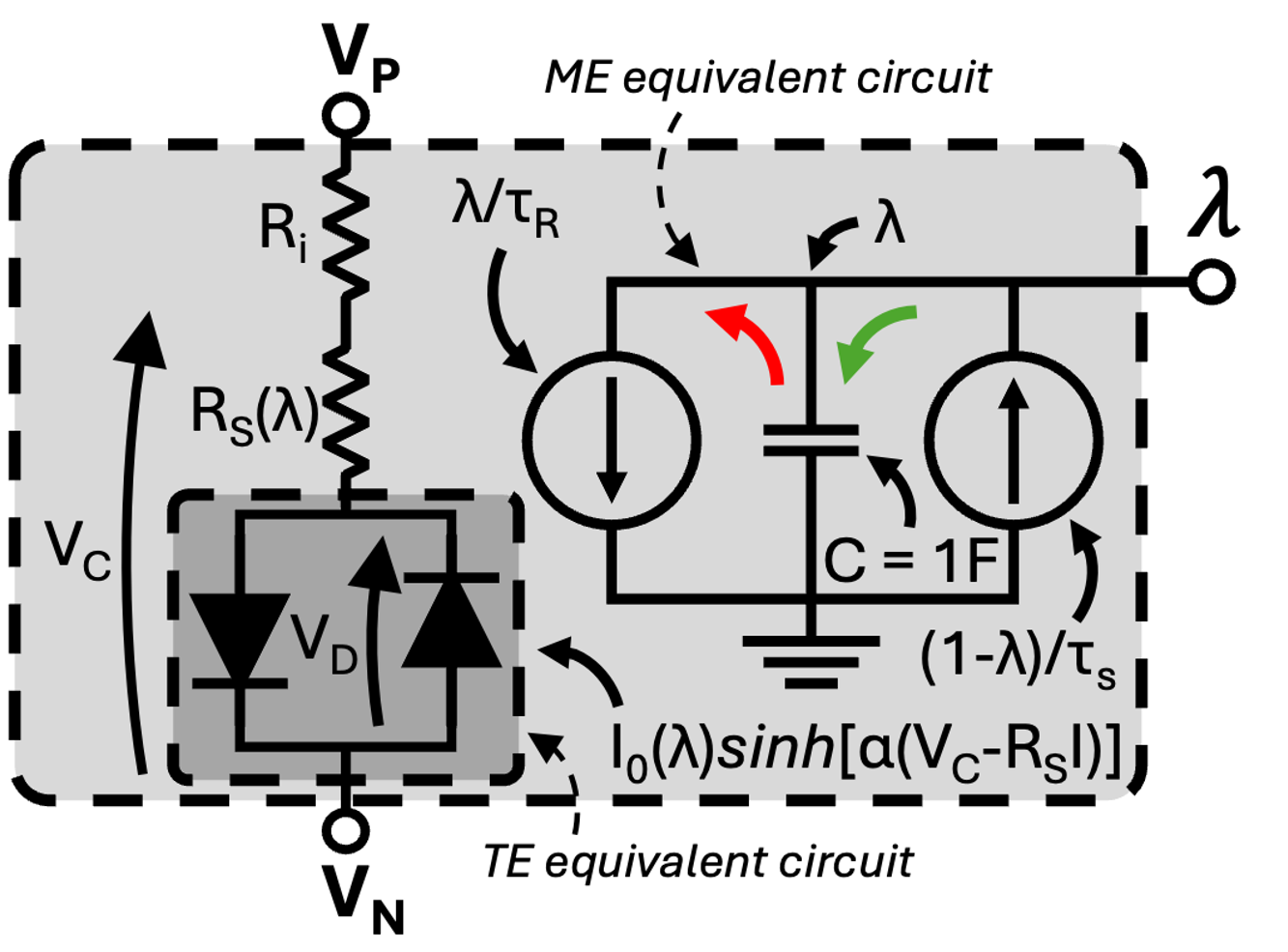}
        \caption{}
        \label{fig:memristor_model}
    \end{subfigure}

\end{minipage}

\caption{(a) Sketch of the typical I/V loop in memristor devices. Nonlinear HRS state conduction is indicated in blue and linear LRS state conduction is indicated in red. Set and reset transitions are indicated to illustrate the transition between the two states. (b) Illustration of the memristor filament's morphology in the HRS and LRS states and the transition between them. Note that in most memristive devices a so-called electroforming step is necessary to allow the device to switch between HRS and LRS states. (c) Equivalent circuit of the dynamic memdiode model employed in the reconfigurable active synapse.}
\label{fig:three_panel_layout}
\end{figure}
\FloatBarrier
\section{References}
\renewcommand{\bibsection}{}

\end{document}